\documentclass[runningheads]{llncs}
\usepackage{times}
\usepackage{epsfig}
\usepackage{graphicx}
\usepackage{subfig}
\usepackage{amsmath}
\usepackage{amssymb}
\usepackage{algorithm}
\usepackage{algorithmic}
\usepackage{amsfonts}
\usepackage{xcolor}
\usepackage{xspace}
\usepackage[pagebackref=true,breaklinks=true,colorlinks,bookmarks=false]{hyperref}
\usepackage{wrapfig}

\def\etal{\mbox{\textit{et al.}}\@\xspace}

\begin{document}
\title{A Unified Batch Selection Policy for Active Metric Learning}
%
%
 \author{Priyadarshini K\inst{1} \and
 Siddhartha Chaudhuri\inst{2} \and
 Vivek Borkar \inst{1} \and
 Subhasis Chaudhuri\inst{1}}
 \institute{IIT Bombay \and Adobe Research\\
 \email{priyadarshini.k@iitb.ac.in,sidch@adobe.com,\{borkar,sc\}@ee.iitb.ac.in}}
\maketitle               

\begin{abstract}

Active metric learning is the problem of incrementally selecting high-utility batches of training data (typically, ordered triplets) to annotate, in order to progressively improve a learned model of a metric over some input domain as rapidly as possible. Standard approaches, which independently assess the informativeness of each triplet in a batch, are susceptible to highly {\em correlated} batches with many redundant triplets and hence low overall utility. While a recent work \cite{kumari2020batch} proposes {\em batch-decorrelation} strategies for metric learning, they rely on ad hoc heuristics to estimate the correlation between two triplets at a time. We present a novel batch active metric learning method that leverages the Maximum Entropy Principle to learn the least biased estimate of triplet distribution for a given set of prior constraints. To avoid redundancy between triplets, our method collectively selects batches with maximum {\em joint entropy}, which simultaneously captures both informativeness {\em and} diversity. We take advantage of the submodularity of the joint entropy function to construct a tractable solution using an efficient greedy algorithm based on Gram-Schmidt orthogonalization that is provably $\left( 1 - \frac{1}{e} \right)$-optimal. Our approach is the first batch active metric learning method to define a unified score that balances informativeness and diversity for an entire batch of triplets. Experiments with several real-world datasets demonstrate that our algorithm is robust, generalizes well to different applications and input modalities, and consistently outperforms the state-of-the-art.


\end{abstract}
\section{Introduction}\label{sec:intro}

\begin{figure}[t!]
\centering
\includegraphics[width=0.48\linewidth]{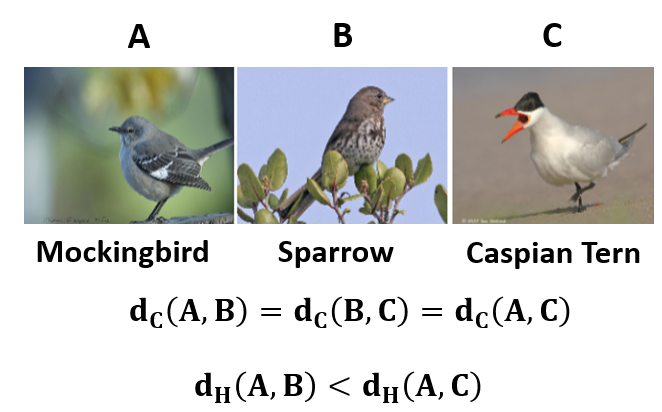}\hspace{2mm}
\includegraphics[width=0.45\linewidth]{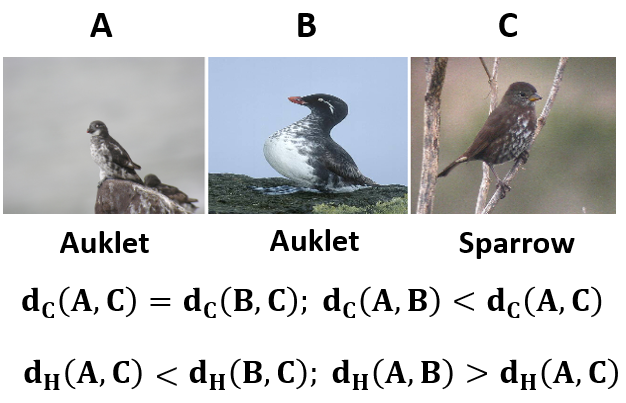}
\caption{Difference between class-based and perceptual distances on two different types of triplets. In each case, the class-based metric $d_C$ fails to capture intra-class variations and inter-class similarities and is not compatible with the perceptual metric $d_H$.}  
\label{F1}
\end{figure}

Understanding similarity between two objects is fundamental to many vision and machine learning tasks, e.g.\ object retrieval~\cite{wang2014learning}, clustering~\cite{xing2003distance} and  classification~\cite{Strese}. Most existing methods model a {\em discrete} measure of similarity based on class labels: all inter-class samples are considered equally dissimilar, even though their features differ by different degrees. But human estimation of perceptual (dis)similarity is often more fine-grained. We may choose, for example, {\em continuous} measures such as the {\em degree} of perceived similarity in taste or visual appearance for comparing two food dishes, rather than discrete categorical labels (Figure \ref{F1}). Thus, it is important to build a {\em continuous} perceptual space to model human-perceived similarity between objects. Recent studies demonstrate the importance of perceptual metrics in several tasks in computer vision and cognitive science~\cite{perceptualMetric,kim2019deep,priyadarshini2019perceptnet}.

\begin{figure}[ht!]
\centering
\includegraphics[width=0.47\linewidth]{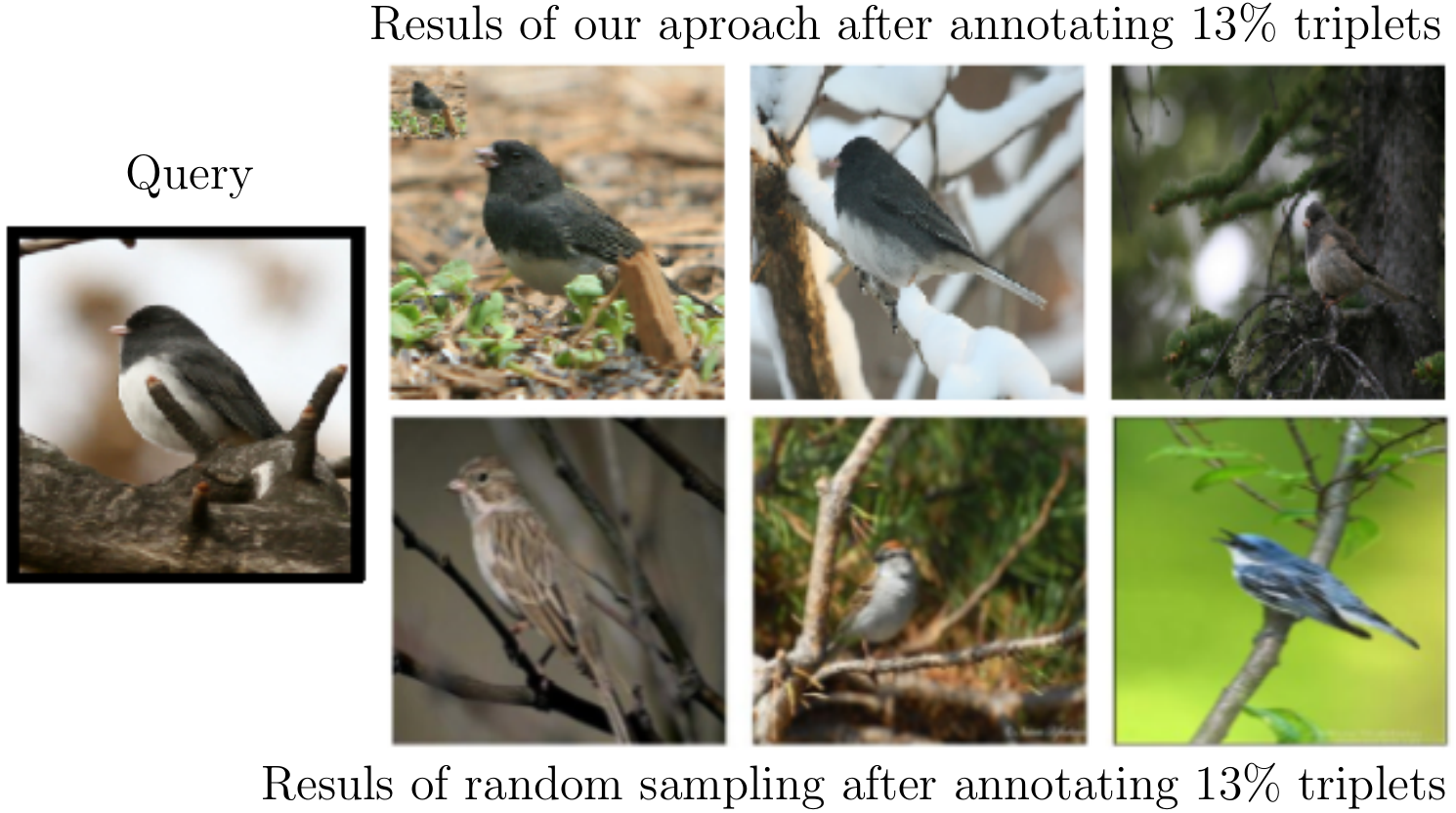}\hspace{3mm}
\includegraphics[width=0.45\linewidth]{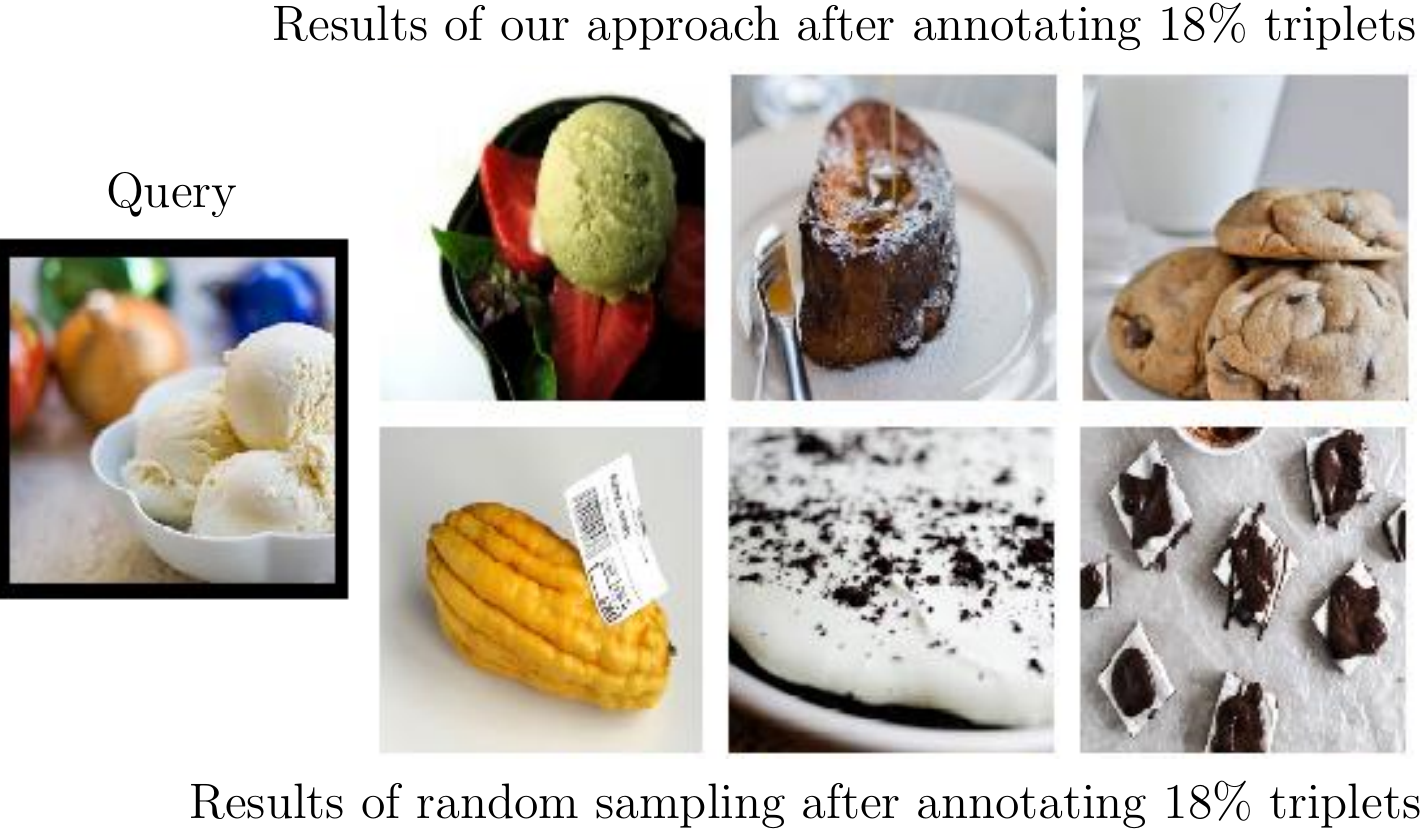}
\caption{Top-3 retrieved images ranked from most to least similar by a perceptual metric (visual appearance for birds, and taste for food) trained on randomly selected (but correctly annotated) triplets vs. high-quality triplets identified for annotation by our method. For a fair comparison, both methods run for equal training rounds and solicit annotations for equal amounts of training data -- 13\% of the CUB-200 bird dataset on the left, and 18\% of the Yummly-Food dataset on the right.}
\label{F2}
\end{figure}

Early work on perceptual metric learning focuses on non-parametric methods (e.g. Multidimensional scaling (MDS)~\cite{kruskal1978multidimensional}) which use numerical measurements of pairwise similarity for training. These are hard to gather and suffer from inconsistency. Instead, similarity {\em comparisons} of the form ``Is object $x_i$ more similar to object $x_j$ than object $x_k$?'' are easier to gather and more stable~\cite{kendall1948rank}. They form a useful foundation for several tasks, including perceptual metric learning. However, the number of possible triplets of $n$ objects is $O(n^3)$, making it infeasible to label even a significant fraction of them. Fortunately, many triplets are redundant and we can effectively model the metric using only a few high-utility triplets (Figure~\ref{F2}). Thus it is imperative to identify and annotate a subset of high-quality triplets that are jointly informative for the model, {\em without knowing the annotations of any triplets in advance.} We stress this last point since it renders common triplet sampling strategies such as (semi-)hard negative mining, which rely on access to a fully annotated dataset, inadmissible.

Active learning is a standard technique that addresses this issue by iteratively identifying small batches of informative samples and soliciting labels for them. While extensively studied for class label-based learning tasks, there exists very little literature~\cite{tamuz2011adaptively,heim2015active,kumari2020batch} on active learning which focuses on perceptual/general metric learning. Further, these works merely assess the informativeness of {\em individual} triplets with uncertainty measures, which assume a triplet with high prediction uncertainty is more crucial to label.  Although effective in many scenarios, such an uncertainty measure makes a myopic decision based solely on the current model's prediction and fails to capture the triplets' collective distribution as a whole. Independently assessed triplets may themselves have much redundancy even if they are individually informative. Hence the triplets should be not merely informative but also {\em diverse} or {\em decorrelated}. 

Kumari \etal~\cite{kumari2020batch} proposed a method for selecting informative {\em and} decorrelated batches of triplets for active metric learning. However, their approach suffers from three major limitations:
(1) The active learning strategy is based on a two stage optimization for informativeness (choice of an overcomplete batchpool of individually informative triplets) and diversity (subsequent trimming of the batchpool), applied sequentially. It does not always ensure an optimal tradeoff between the two criteria.
(2) The proposed diversity measures are all ad-hoc with no principled connection to informativeness. Being heuristic, no single measure works consistently well in all cases, making it harder for a user to select which measure to use in practice.
(3) The informativeness of a triplet is determined using a point estimate of the perceptual metric. Bias in the latter, e.g., because of suboptimal batch selection in prior iterations, directly translates to bias in informativeness, which can misguide the strategy.

To mitigate these issues, we propose a new batch active learning algorithm developed specifically for triplet-based metric learning. Our {\bf key insight} is to express a set of (unannotated) triplets as a vector of random variables, and select batches of triplets that maximize the {\em joint entropy} measure.  Thus, instead of separately expressing and optimizing informativeness for individual triplets and diversity for pairs of triplets, we develop a single probabilistic informativeness measure {\em for a batch of triplets}. We also provide computationally efficient approximate solutions with provable guarantees. Specifically, our main technical contributions are:

\begin{enumerate}
\item We propose to use the joint entropy of the distribution of triplet margins to rank a batch of unannotated triplets. We estimate the second-order statistics (mean and covariance) of {\em triplet margins} by randomly perturbing the current model trained on prior batches as in~\cite{gal2016dropout}, to characterize the distribution.

\item Using the Maximum Entropy Principle, we arrive at a Gaussian distribution compatible with  the given empirical mean and covariance, whose entropy  is characterized by the determinant of the covariance matrix. As exact maximization of the joint entropy is prohibitively expensive (there are $\binom{m}{b}$ possible batches of size $b$ from $m$ triplets), we use the fact that entropy is monotone increasing and submodular to justify a greedy policy which is provably \mbox{$(1 - \frac{1}{e})$}-optimal~\cite{nemhauser1978analysis}.

\item We achieve further computational efficiency by using the fact that the covariance matrix is a Gram matrix, and its determinant can be computed using efficient recursion. Our method recursively maximizes successive projection errors of a set of vectors, picked one at a time, when projected onto the span of previous choices. This amounts to successive  maximization of the conditional entropy, and is easily implemented using Gram-Schmidt orthogonalization.
\end{enumerate}

We demonstrate the effectiveness of our approach through extensive experiments on different applications and data in different modalities (image, taste and haptic). In addition to having a sound theoretical justification, our method provides a significant performance gain over the current state-of-the-art.

\section{Related Work} \label{sec:related}

The prior work can be roughly divided into three categories. We review representative techniques in each and discuss how our work  differs from the existing methods.

\subsection{Perceptual Metric Learning}

While there is extensive recent research on distance metric learning, most of the algorithms are specific to class-based learning tasks such as classification \cite{Strese} and clustering \cite{xing2003distance}, which consider two objects similar if they belong to the same class. See Bellet \etal~\cite{bellet2013survey} for a comprehensive review. In contrast, our goal is to define a perceptual distance that captures the degree of similarity between any two objects irrespective of their classes. Recently, a whole new literature has emerged that emphasizes the importance of learning such continuous measures of similarity for various applications, e.g.\ for measuring image similarity \cite{perceptualMetric}, face recognition \cite{fan2019scoot}, concept learning \cite{wang2014learning,wah2015learning} and perceptual embedding of objects \cite{priyadarshini2019perceptnet,jamieson2011low,GNMDS}.
The closest application to ours is perceptual embedding of objects, where the embedding function is learned so as to model the human-perceived inter-object similarity. While multidimensional scaling (MDS) techniques have been extensively applied for this \cite{jamieson2011low,GNMDS,kruskal1978multidimensional}, they are non-parametric and require numerical similarity measurement as inputs, which are hard to gather \cite{kendall1948rank}. Recent works \cite{perceptualMetric,mcfee2011learning} address these limitations by developing parametric models using non-numeric relative comparisons. A relevant method is the triplet-based deep metric learning method of Kumari \etal~\cite{priyadarshini2019perceptnet}. Although our method borrows base metric learning architectures from \cite{priyadarshini2019perceptnet},\cite{priyadarshini2019perceptnet}  doesn't aim to make the metric learning algorithm data-efficient by developing an active data sampling technique.

\subsection{Active Learning for Classification}

Active learning (AL) methods have been well explored for vision and learning tasks, see Settles \cite{settles2012} for a detailed review of active learning methods for class-based learning. Typically, the AL methods select a single instance with the maximum individual utility for annotation in each iteration. The utility of an instance is decided by different heuristics, e.g.\ uncertainty sampling \cite{kumari2020batch}, query-by-committee (QBC) \cite{gilad2006query}, expected gradient length (EGL) \cite{ash2019deep}, and model-output-change (MOC) \cite{freytag2014selecting}. The simplest and most widely applicable uncertainty sampling approach has  been extended to modern deep learning frameworks and variational inference~\cite{sinha2019variational}. However, in all these methods, each sample's utility is evaluated independently without considering dependence between them.

In batch-mode active learning, data items are assessed not one at a time but in batches, to reduce the number of times the model is retrained. To avoid selecting correlated batches, some recent attempts evaluate the whole batch's utility by taking mutual information between samples into account. In contrast to our work, most of them are developed for classification tasks \cite{ash2019deep,kirsch2019batchbald,CNN2018ICLR,pinsler2019bayesian}. For example, Kirch \etal~\cite{kirsch2019batchbald} define the utility score as the mutual information between data points in a batch and model parameters and then pick a subset with the maximum score. Pinsler \etal~\cite{pinsler2019bayesian} formulate the active learning problem as a sparse subset selection problem approximating the complete data posterior of the model parameters. Both methods have a similar motivation to our work, but they are developed for the classification task, and their informativeness measures are not easy to extend to the metric learning task. Ash \etal~\cite{ash2019deep} use the norm of the gradient at a sample to implicitely capture both informativeness and diversity, and select a subset of the farthest samples in the gradient space. This ensures both informativeness and diversity by a single gradient-based measure, which does not work well in the metric learning task, as shown by Kumari \etal~\cite{kumari2020batch}. Sener and Savarese~\cite{CNN2018ICLR} follow a similar strategy in a different feature space. Shui \etal~\cite{shui2020deep} introduce a unified approach for training and batch selection process and explicitly define uncertainty-diversity trade-off by adopting Wasserstein distance. 

\subsection{Active Learning of Perceptual Metrics}

There are only a few works on active learning of a perceptual metric. Most of these, e.g. \cite{tamuz2011adaptively,heim2015active}, are based on a single instance evaluation criterion. They define the utility of a single triplet and select a batch of the individually highest-utility triplets to annotate. In contrast, we define a utility score for a batch taking joint information between triplets into account. The closest work to ours is a very recent paper by Kumari \etal~\cite{kumari2020batch}. The algorithm involves a two-stage process. First, it selects an overcomplete set of individually highly informative samples, and then subsamples a less correlated subset, using different triplet-based decorrelation heuristics, as the current batch. This method, in essence, is still based on a single triplet selection strategy. In contrast, we present a new, rigorous approach to define utility for a batch as a whole based on {\em joint entropy}, providing a unified utility function to balance both informativeness and diversity.

\section{Proposed Method}\label{sec:method}

In this section, we first briefly describe the perceptual metric learning setup and the underlying neural network-based learner called PerceptNet~\cite{priyadarshini2019perceptnet}. Next, we introduce our novel batch selection policy explicitly designed for triplet-based active metric learning.

\subsection{Triplet-Based Active Metric Learning}\label{sec:metric_learning}

Let $X = \{x_i\}_1^n$ represent a set of $n$ objects, each described by a $d$-D feature vector $x_i$. Also, let $T_L$ be a set of ordered triplets, where each triplet $(x_i, x_j, x_k)$ indicates that the object $x_i$ is more similar to object $x_j$ than to $x_k$. For brevity we denote $(x_i, x_j, x_k)$ by ${ijk}$. We frame the perceptual metric learning problem as learning an embedding $\phi:R^d \rightarrow R^{\hat{d}}$, s.t. the $L_2$ distance between any two objects in the embedding space $d_{\phi}(x, y) = \left\|\phi(x) - \phi(y)\right\|$ reflects the perceptual distance between them. In recent work, $\phi$ is typically modeled with a neural network: in our experiments, we choose the existing PerceptNet model~\cite{priyadarshini2019perceptnet}, where three copies of the same network, with shared weights, process three objects $x_i$, $x_j$ and $x_k$ during training. The output is optimized with an exponential triplet loss $\mathcal{L} = \sum_{T_L} 
 e^{-\left(d^2_{\phi}(x_i, x_k) - d^2_{\phi}(x_i, x_j) \right)}$ to maximize the distance margins (a.k.a ``triplet margins''), as defined by the exponent, for training triplets.

The number of possible triplets is cubic in the number of objects, so annotating a significant fraction of them is often intractable, e.g.\ in domains such as haptics and food tasting where annotation is especially slow. However, an effective embedding can be modeled with far fewer comparisons if triplets are sampled selectively based on {\em how much information} they would provide if annotated. This calls for active learning. The model is trained iteratively: batches of triplets informative to the current model are selected for annotation in each iteration, after which the model is retrained. However, the efficiency gain of selecting larger batches may be undone by {\em correlation} among triplets in a batch implying low overall information, a common issue in independent optimization of  individual informativeness of each  triplet. To mitigate this, prior works have studied {\em batch decorrelation} strategies for classification~\cite{kirsch2019batchbald,ash2019deep,pinsler2019bayesian}. Recently, Kumari \etal~\cite{kumari2020batch} developed a decorrelation strategy for metric learning with separate steps for optimizing individual triplet informativeness and then batch diversity. However, as already noted, this work suffers from limitations related to their design choices. In contrast, we develop a method that jointly defines and optimizes the informativeness of an entire batch while implicitly ensuring diversity. The method is grounded in the Maximum Entropy Principle and leads to an attractive computational scheme.

\subsection{Joint Entropy Measure for Batch Selection}\label{sec:batch_AL}

The key to a good batch mode active learning is an effective informativeness measure for a batch of triplets. For tractability, earlier work typically defines a measure adding up the individual informativeness scores of triplets. A popular score is the Shannon entropy of the prediction probability $p_{y}$ of the current model trained on prior batches, for a triplet $t$ taking one of two possible orderings $y \in \{ijk, ikj\}$, $H(t) = -\sum_{y \in \{ijk, ikj\}}p_y \log p_y$~\cite{tamuz2011adaptively}.
While often termed ``uncertainty'', this is not a good predictor of actual model uncertainty due to possible bias in the current model~\cite{gal2016dropout}. Further, individually high-entropy triplets may also have high mutual information, hence simply adding up the scores may overestimate the actual utility of the batch.

We propose a novel batch selection algorithm based on the {\em joint entropy} of an entire batch of triplets $B$, capturing their mutual dependence. We define the joint probability distribution of a set of unannotated triplets on some feature space such as their distance margins. This probability is defined using the {\em distribution of likely models} given prior batches, reducing any bias due to model training. Note that this is {\bf quite different} from the {\em prediction probability} of a single fixed model, described above. The joint distribution over a set of triplets naturally captures the notion of interdependence among them.

\subsection{Maximum-Entropy Model of the Joint Distribution}\label{sec:maxent}

Our goal is to postulate the joint probability distribution of unannotated triplets in a batch, preferably in a form that allows efficient computation of its entropy. We represent each triplet $t$ by its distance margin \mbox{$\xi_t = d^2_{\phi}(x_i, x_k) - d^2_{\phi}(x_i, x_j)$}. Then a batch, denoted $B = \{t_1, t_2, \dots, t_b\}$ is represented by the vector of distance margins given by \mbox{$\vec{\xi}_B = [\xi_{t_1}, \xi_{t_2}, \dots, \xi_{t_b}]$}.\footnote{Other triplet based representations are possible: we found the above to be a consistent and more useful feature in practice.} We assume there is uncertainty about these margin predictions arising from the fact that there is a {\em distribution of plausible models} given the previously annotated data. Hence, each distance margin $\xi_{t_i}$ is a 1D random variable taking different values for different choices of model parameters $\phi$. As discussed above, simply looking at the predicted ordering probabilities of individual triplets is both error-prone and fails to consider correlation between triplets. Fortunately, if the model is a neural network, it has been shown that random dropout yields a good Bayesian approximation of model uncertainty~\cite{gal2016dropout}. We stochastically apply the dropout $K$ times to the model, evaluating $\vec{\xi}_B$ each time, to sample the joint margin vector distribution of the batch and to compute the corresponding $b$-dimensional mean and covariance matrix. We invoke the Maximum Entropy Principle~\cite{Jaynes} which maximizes the Shannon entropy subject to constraints on prescribed averages. The maximum entropy distribution, consistent with all prior constraints, ensures the largest amount of uncertainty with respect to unknown, and hence introduces no additional biases in the estimation. Empirical estimates of the entropy of the batch from samples are susceptible to noise, and lead to a hard combinatorial optimization over batches. So  we constrain the mean and covariance matrix of the triplet margins to match their empirical values $\vec{\mu}_B$ and $\Sigma_B$ and maximize the differential entropy $H(B) = -\int p(\vec{\xi}_B) \log p(\vec{\xi}_B)d\vec{\xi}_B$ subject to these constraints. This leads to a multivariate gaussian distribution $N(\vec{\mu}_B, \Sigma_B)$ with entropy
\begin{equation}\label{joint_entropy}
\begin{aligned}
 H(B) = \frac{1}{2}\log \left( (2\pi e)^{b} \det(\Sigma_B) \right)
\end{aligned}
\end{equation}
Note that this score takes into account inter-triplet correlation, unlike measures depending only on individual marginals. The next task is to efficiently select an optimum batch of size $b$ with maximum informativeness: \mbox{$B^* = \arg \max_{B \subset T_U, |B| = b} H(B)$}, where $T_U$ is the set of currently unannotated triplets.

\subsection{Greedy Algorithm for Batch Selection}

Since the maximization of the joint entropy function $H(B)$ over subsets is computationally prohibitive, we use the fact that entropy is monotone increasing and submodular to justify a greedy policy which is provably $(1 - \frac{1}{e})$-optimal by the results of Nemhauser \etal \cite{nemhauser1978analysis}.
The greedy algorithm builds up the set $B^*$ incrementally. In step $k$, we pick the triplet $t_k$ which has maximum conditional entropy given triplets $B_{k - 1}$ selected in previous steps. Specifically,
\begin{eqnarray}\label{conditional_entropy}
\lefteqn{t_k  =  \arg \max_{t \in T_U \setminus B_{k - 1}} H( \{ t \} \mid B_{k - 1})} \nonumber \\
& = & \arg \max_{t \in T_U \setminus B_{k - 1}} H(B_{k - 1} \cup \{ t \}) - H(B_{k - 1}) \nonumber \\
& = & \arg \max_{t \in T_U \setminus B_{k - 1}} \log \left( \frac{ \det \left( \Sigma_{B_{k - 1} \cup \{ t \}} \right) }{ \det \left( \Sigma_{B_{k - 1}} \right )} \right).
\end{eqnarray}
This step is repeated $|T_U|$ times. The greedy policy has low complexity (quantified later) and scales well to large datasets. The overall batch selection algorithm is listed in \ref{alg:algorithm}. The remaining challenge is to efficiently compute the increment in the determinant of the covariance matrix in each step. We present a recursive algorithm for this, which also clarifies why the method selects a decorrelated batch.

\begin{algorithm}[!t]
\caption{Greedy algorithm to maximize $H(B)$}
\label{alg:algorithm}
\textbf{Input}: Unlabeled triplets $T_U$, batch size $b$, entropy function $H:2^{T_U} \rightarrow \mathbb{R}$ as in Eq. \ref{joint_entropy}. \\
\textbf{Output}: Batch $B$ that is an $(1 - \frac{1}{e})$-approximation to $\arg \max_{B \subset T_U, |B| = b} H(B)$.\\
\begin{algorithmic}[1] 
\STATE $B_0 \leftarrow \emptyset $, $H(B_{0}) = 0$
\FOR{k=1,\dots, b}
\STATE $t_k \leftarrow \arg \max_{t \in T_U \setminus B_{k-1}} \log \left({\det \left( \Sigma_{B_{k - 1} \cup \{ t \}} \right)}/{\det \left(\Sigma_{B_{k - 1}}\right)}\right)$  \ \ \  \STATE $ B_k \leftarrow B_{k-1} \cup \{ t_k \} $
\ENDFOR
\STATE \textbf{return} Final subset $B_k$
\end{algorithmic}
\end{algorithm}

\subsection{Recursive Computation of Determinant of Covariance Matrix}\label{sec:recursive_det}

The covariance matrix is a Gram matrix, i.e. its $(i,j)^\text{th}$ element can be written as the dot product of the $i^\text{th}$ and $j^\text{th}$ vectors from a given family of vectors. This allows us to recursively compute its determinant and choose the recursion order according to the greedy policy for approximate optimization. Let $u_t$ denote the zero-mean vector of all sampled distance margins, $[\xi_t(\phi_1), \cdots, \xi_t(\phi_K)] - [\mu_{t}, \cdots, \mu_t]$, for a single triplet $t$. The covariance matrix $\Sigma_{B_{k - 1}}$ has the form $UU^T$, where each column of $U$ is $u_s, s \in B_{k - 1}$. In the $k^\text{th}$ step, a new row and column vector for a new triplet $t$ are appended to $U$. Using the Gram matrix property, we have $\det(\Sigma_{B_{k - 1} \cup \{ t \}}) - \det(\Sigma_{B_{k - 1}}) = \|\tilde{u}_k\|^2$, where $\tilde{u}_k$ is the normal from $u_t$ onto $\text{span}\{u_s \mid s \in B_{k - 1}\}$. Thus the scheme successively maximizes the squared projection error $\|\tilde{u}_k\|^2$, over the remaining vectors $\{u_t \mid t \in T_U \setminus B_{k-1}\}$. Thus we select at each step the triplet that is least correlated with the already chosen triplets. The orthogonal projections are  computed using the modified Gram-Schmidt orthogonalization scheme from \cite{hoffmann1989iterative}, with complexity $dn^2$, where $d$ is the dimension of the ambient vector space and $n$ the number of vectors. Since we compute the projection error for all $|T_U| - n \approx |T_U|$ remaining triplets at each step (because $|T_U| >> n$), the overall complexity of the scheme is $dn^2|T_U|$. 

In summary, the submodularity of the joint entropy function naturally combines informativeness, diversity, and representativeness, which are precisely the desired properties for batch mode active learning.
\section{Experiments}\label{sec:results}

We perform several experiments to answer the following questions: (1) Is our method competitive with standard baselines, including the state-of-the-art method(s), for different choices of hyperparameters, feature dimension, applications, and datasets? (2) How good is our assumption that the second-order statistics (mean and covariance) are sufficient statistics for estimating the reasonable distribution? (3) How robust is our method to labeling error? 
We address these questions by conducting several experiments on real-world datasets with different modalities: image, food and haptic. For each of these datasets, we select an appropriate neural network architecture -- for the haptic and food datasets we ensure that these architectures exactly match those of Kumari et al.~\cite{kumari2020batch} so that the comparison is fair (\cite{kumari2020batch} did not present any result on images, requiring us to implement their method on image databases). We test with different initial pools and varying batch sizes. We also simulated random errors in the triplet orderings to test robustness to labeling error. 

\paragraph{\textbf{Datasets}.} We evaluate the performance of our method on five real-world datasets for which triplets defining perceptual metrics are available: Yummly food dataset~\cite{wilber2014cost}; TUM haptic texture dataset~\cite{Strese}; Abstract500 image dataset~\cite{robb2015crowdsourced}; CUB-200 image dataset~\cite{wah2015learning}, and Scoot facial sketch dataset~\cite{fan2019scoot}. The {\bf Yummly-Food} dataset has 72148 triplets defined over 73 food items based on taste similarity. Each food item is represented by a 6D feature vector (this is an experiment with a low feature dimension) with each component indicating different taste properties. We use 20K training and 20K test triplets sampled from the entire set of triplets. The {\bf TUM-Haptic} dataset contains signals from 108 different types of surface materials. Each type of material has 32-D spectral feature vectors for 10 representative acceleration traces. The triplets are generated from a given ground-truth perceptual matrix, which has user-recorded perceived similarity responses. Like the Yummly-Food dataset, we have training and test sets of 20K triplets each. We also evaluate our method on a comparatively larger dataset (but relatively small for image data), the {\bf Abstract500} image dataset~\cite{robb2015crowdsourced}, which contains 500 images of $128 \times 128$ pixels, with pairwise perceptual similarities between them. Each image is represented by a 512-D GIST feature (an example of a relatively high-dimensional feature vector) extracted using 32 Gabor filters at four scales and eight orientations~\cite{oliva2001modeling}. We use perceptual matrix to generate 20K training and 20K test triplets. Next, we use the popular and much larger {\bf CUB-200} bird database that contains 200 bird species with roughly 30 images in each class. We choose five representative images for each class and generate its features using a pretrained ResNeXt-101-32x8d model. The network takes segmented images as input and outputs \mbox{2048-D} feature vectors. The training and test sets each have 10K triplets sampled from the entire set of 93530 triplets. Finally, the results on the {\bf Scoot} dataset, which is relatively small, consisting of just 1282 triplets, are presented in the supplementary material because of space constraints.

\paragraph{\textbf{Baselines}.} We compare our method with five baselines, including the state-of-the-art method: (1) \mbox{\bf US-$\langle$Dist$\rangle$:} A batch of individually high-entropy triplets is pruned subjected to different (denoted by $\langle$Dist$\rangle$) decorrelation measures to select a diverse batch of informative triplets~\cite{kumari2020batch}. It is the current state-of-the-art for batch mode active metric learning, and outperforms other alternatives like BADGE~\cite{ash2019deep} (adapted to metric learning). We pick $\langle$Dist$\rangle$ to be the highest-performing variant in each individual experiment. (2) \mbox{\bf Variance:} Triplets with the highest individual distance-margin variance across a collection of models generated using dropout \cite{kendall2015bayesian}. This method simulates the effect of replacing the joint entropy of a batch with the sum of individual entropies of triplets in the batch. (3) \mbox{\bf Random:} A passive learning strategy that uniformly samples each batch of triplets at random. Though na\"{i}ve, this choice often results in reasonably good accuracy. (4) \mbox{\bf US:} Uncertainty method, which picks the top $b$ triplets with highest uncertainty in predicted triplet ordering (i.e. the model's (lack of) ordering confidence), without taking correlation among them into account~\cite{tamuz2011adaptively}. (5) \mbox{\bf BADGE:} A diverse set of triplets with maximum loss gradients for the most probable label, selected using k-means~(\cite{ash2019deep} adapted to the triplet scenario).

\begin{figure*}[!t]
\centering
\vspace{-3mm}
\begin{tabular}{ccc}
\includegraphics[width=0.327\linewidth]{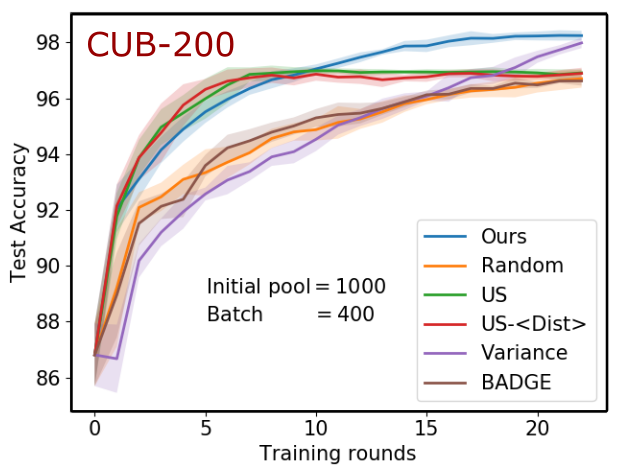}
\includegraphics[width=0.327\linewidth]{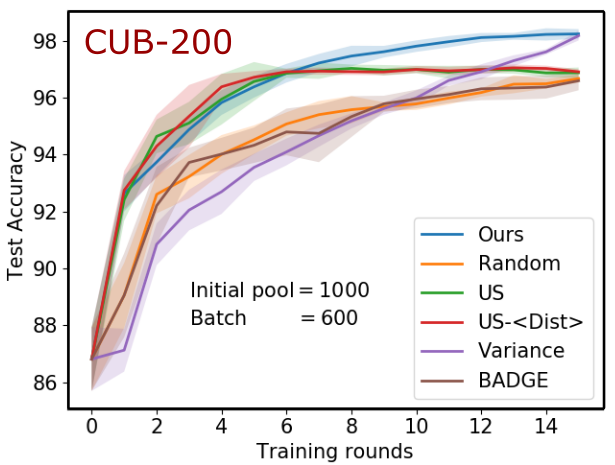}
\includegraphics[width=0.327\linewidth]{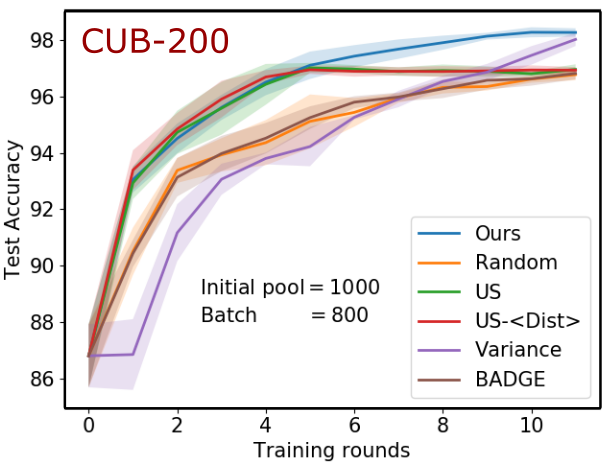}\\
\includegraphics[width=0.327\linewidth]{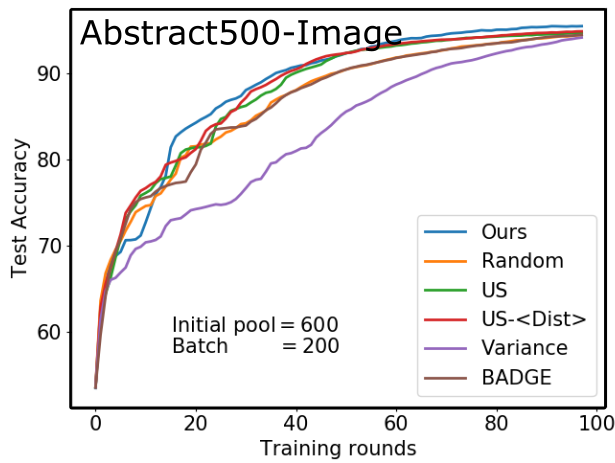}
\includegraphics[width=0.327\linewidth]{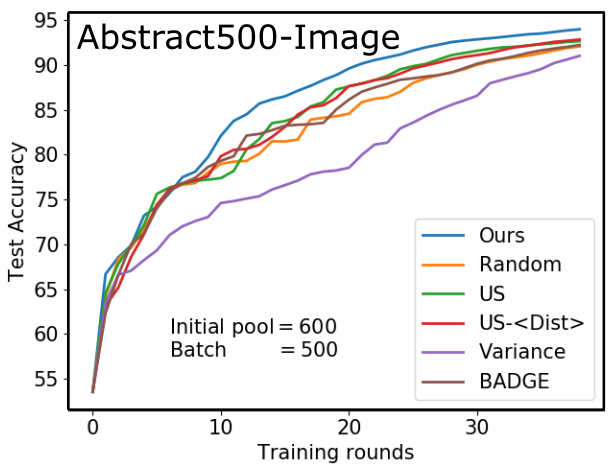}
\includegraphics[width=0.327\linewidth]{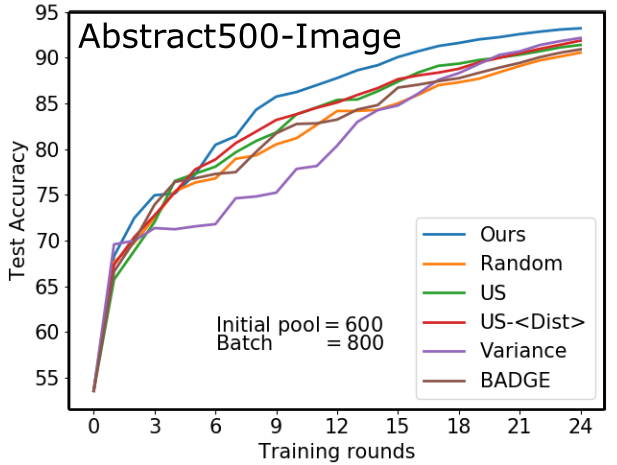}
\end{tabular}
\caption{Performance of different active learning methods on two image datasets CUB-200 and Abstract500 for increasing batch sizes (400/600/800 or 200/500/800, from left to right). Here accuracy means what fraction of test triplets have been detected with the correct ordering. To avoid clutter, standard deviations are shown only for the CUB-200 dataset and the rest are shown in the supplementary material.\vspace{-5mm}}
\label{Q0}
\end{figure*}

\paragraph{\textbf{Active Learning Setup}.} For the CUB-200 dataset, we begin each experiment with an initial pool of 1000 annotated triplets, and for the other three datasets with 600 annotated triplets, and pretrain the model $\phi_0$, which is used as a common starting point for all compared methods. In each active learning iteration, we select the best fixed-size batch of unannotated triplets, using the chosen batch selection method, and acquire their orderings. To make convergence faster, we update the current model $\phi_i$ to obtain $\phi_{i+1}$ using the available  additional annotated triplets instead of training \textit{ab initio}. The performance of the learned model is evaluated by its \textit{triplet generalization accuracy}, which denotes the fraction of triplets whose ordering is correctly predicted~\cite{priyadarshini2019perceptnet}. Each experiment is repeated with five random train/test splits, and the average performance along with the standard deviation is reported. (For most plots, the standard deviation is shown in supplementary material, for clarity.)

\paragraph{\textbf{Implementation Details}.} The architecture and training hyperparameters used for different datasets are as follows: Yummly-Food: 3 fully-connected (FC) layers with 6, 12 and 12 neurons; TUM-Haptic: 4 FC layers with 32, 32, 64 and 32 neurons; Abstract500-Image: 6 FC layers with 512, 256, 128, 64, 32 and 16 neurons; CUB-200: 3 FC layers with 2048, 512 and 32 neurons. Each layer is followed by a dropout layer with a dropout probability of 0.02. The Adam optimizer~\cite{kingma2014adam} is used for training all models with a learning rate of $10^{-4}$ for Yummly-Food, TUM-Haptic, and Abstract500-Image dataset, and $10^{-5}$ for CUB-200 dataset. The model is trained with an SGD batch size of 500 for 1000 epochs for all four datasets.

\begin{figure*}[!t]
\centering
\vspace{-3mm}
\begin{tabular}{ccc}
\includegraphics[width=0.327\linewidth]{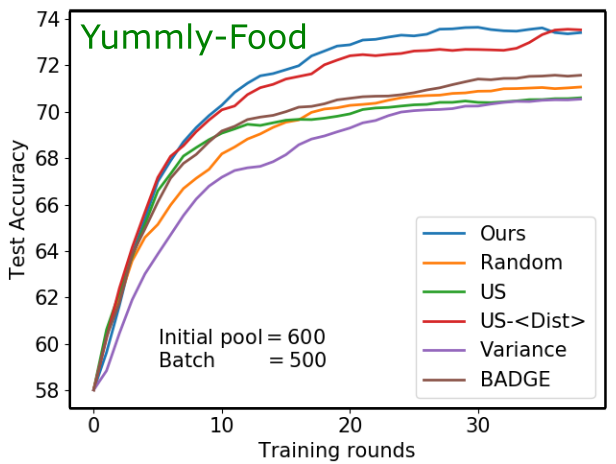}
\includegraphics[width=0.327\linewidth]{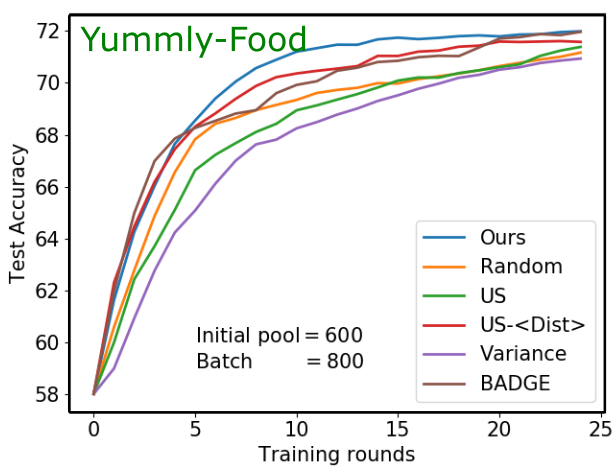}
\includegraphics[width=0.327\linewidth]{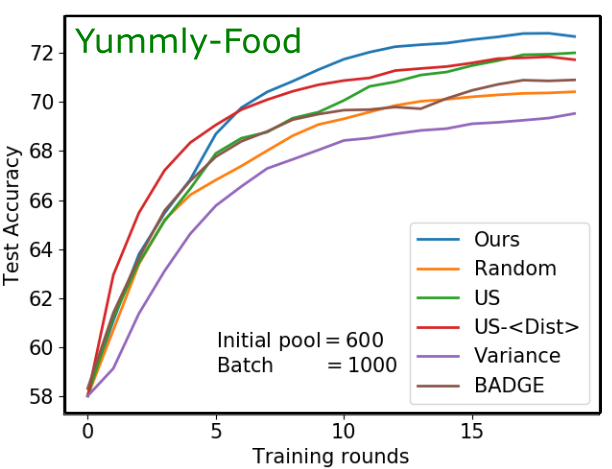}\\
\includegraphics[width=0.327\linewidth]{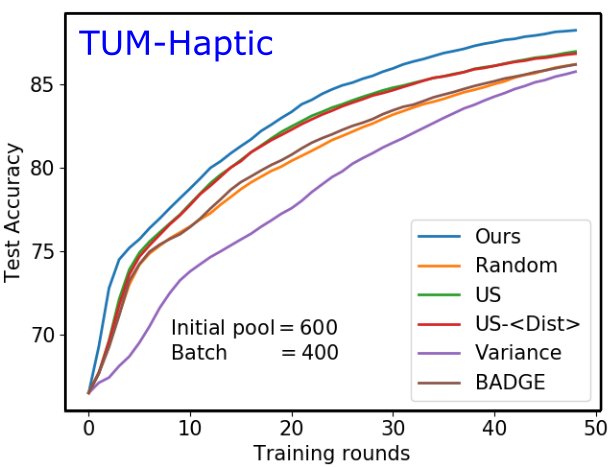}
\includegraphics[width=0.327\linewidth]{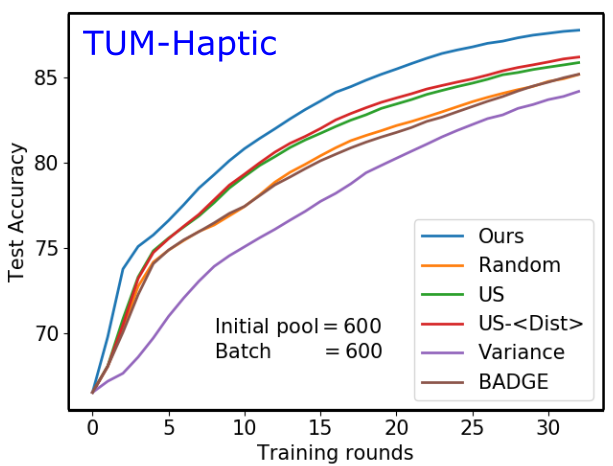}
\includegraphics[width=0.327\linewidth]{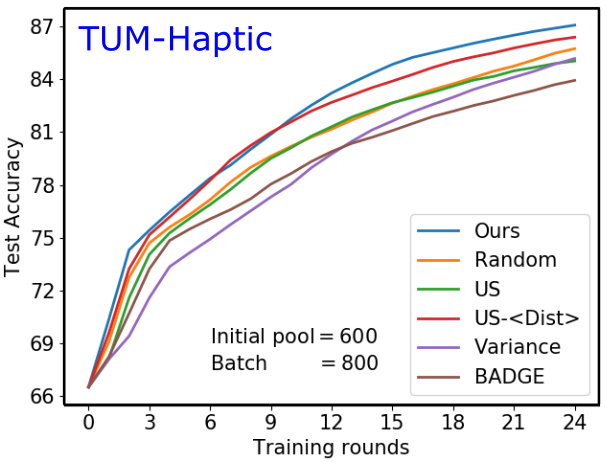}
\end{tabular}
\caption{Performance of different active learning methods on Yummly-Food and TUM-Haptic texture datasets for varying batch sizes (500/800/1000 or 400/600/800).\vspace{-5mm}}
\label{Q1}
\end{figure*}

\paragraph{\textbf{Active Learning Performance}.} The performance of our method against the baselines described earlier is plotted in Figure~\ref{Q0} for the image datasets CUB-200 and Abstract500-Image, shown as a function of the number of active learning iterations. In Figure~\ref{Q1}, we compare the performances of all methods for the data from other modalities, i.e., haptic and food. We observe that our method is consistently better than the state-of-the-art \mbox{\bf US-$\langle$Dist$\rangle$} method (for clarity, we only show the specific variant offering the best performance in each experiment). Our method reaches higher accuracies quicker and also tends to converge to a higher final accuracy on both Yummly-Food and TUM-Haptic datasets.
For the large CUB-200 image dataset, our method is neck-and-neck with the state-of-the-art for the first few iterations and then rapidly overtakes it, widening the gap with additional iterations. For the smaller Abstract500 image dataset, the improvements are more prominent with larger batch sizes, reflecting the focus of our work on batch-mode learning. Additionally, for the CUB-200 dataset, we plot the standard deviation in the same plot as the shaded region (of the same color) around the performance curves for different methods (standard deviations on other datasets are shown in supplementary). Even though the figure looks a little cluttered, one can see that the standard deviation for the proposed method is better than that of the next-best method, signifying a more consistent performance. This substantiates our claim that joint entropy is a better batch score than an ad-hoc combination of independent informativeness and diversity heuristics. Further, our method does not require the user to select a suitable decorrelation heuristic to manually fine-tune the performance.

\begin{figure}[!t]
\centering
\vspace{-3mm}
\includegraphics[width=1\linewidth]{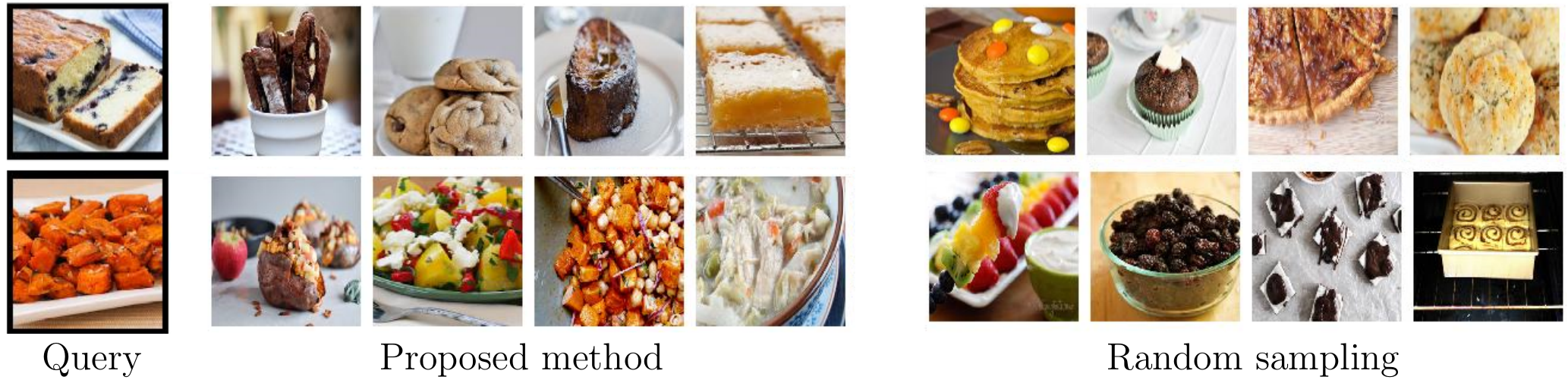}\\
\includegraphics[width=1\linewidth]{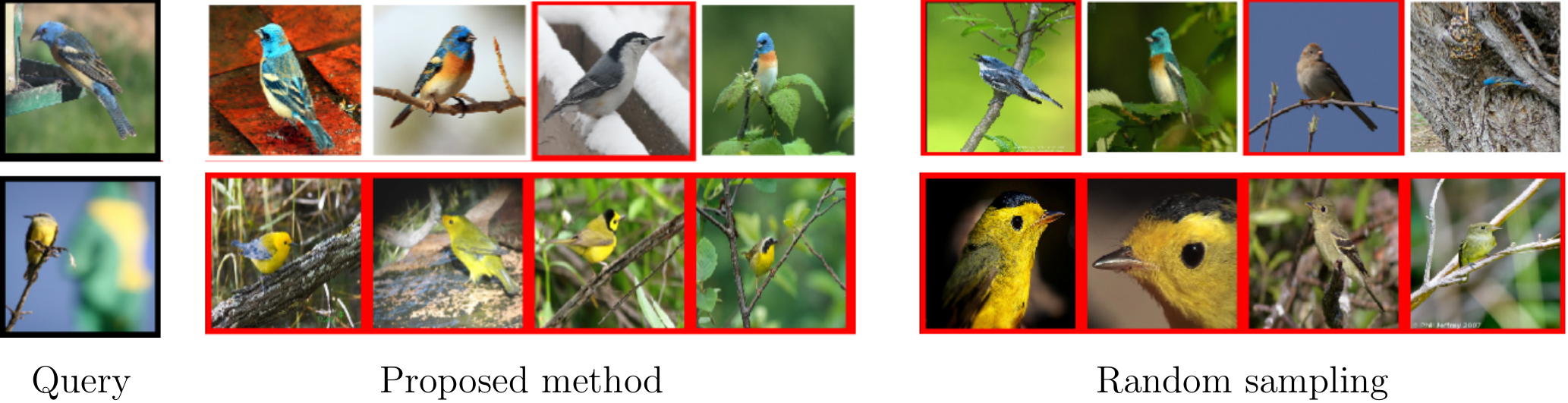}
\caption{Top-4 retrieved images in the order of increasing perceptual distance, left to right, using our method and random sampling (randomly-selected batches are annotated for training) on different modalities datasets. On both datasets, each model is trained for twelve training rounds, constituting 18\% of training triplets. Images different from the query class are bounded by a red box, substantiating that two images from different classes can be perceptually more similar than two from the same class. The {\em triplet order accuracy}, defined here as the number of test triplets whose order is preserved by the ranked list of retrieved images, for our method vs random sampling after the 12$^\text{th}$ round of training is $M^{12}_{\text{Ours}} = 96.7\%$, $M^{12}_{\text{Random}} = 92\%$ for image dataset and $M^{12}_{\text{Ours}} = 72.9\%$, $M^{12}_{\text{Random}} = 69.3\%$ for food dataset. More results with different queries and learned metrics are shown in the supplementary material.\vspace{-5mm}}
\label{Q2}
\end{figure}

We also outperform the other two baselines: {\bf Random} and {\bf Variance}. It is particularly informative to see the generally poor performance of {\bf Variance} (lower than the Random). Because of high correlations among informative triplets, individually selecting the most informative triplets does not learn the entire metric space as well as just picking triplets at random. In contrast, our method as well as that of Kumari \etal \cite{kumari2020batch} both incorporate batch decorrelation and outperform random sampling. This shows the critical importance of batch diversity in an active learning strategy.

Next, we evaluate the effectiveness of our method for an {\bf object retrieval} task. Specifically, we compare our method with the random sampling baseline at different training rounds. We show the retrieval results on two different modalities, food and image. We split the Yummly-Food dataset into 40000 training and 32148 test triplets, and the CUB-200 dataset into 40000 training and 33000 test triplets. On both datasets, we perform active learning with a batch size of 600 and an initial pool of 500 triplets. For a given query image, the top four instances from the retrieval set are shown (ranked from most similar to least similar) in Figure~\ref{Q2}. As we can see, retrieval results of our method resemble the query in taste or visual appearance better than the random sampling. Please see the supplementary material for further results from this experiment.

\begin{figure}[!t]
\centering
\vspace{-3mm}
\includegraphics[width=0.34\linewidth]{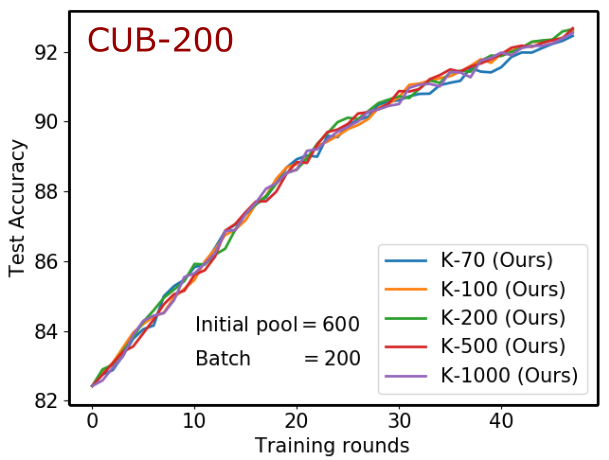}
\includegraphics[width=0.32\linewidth]{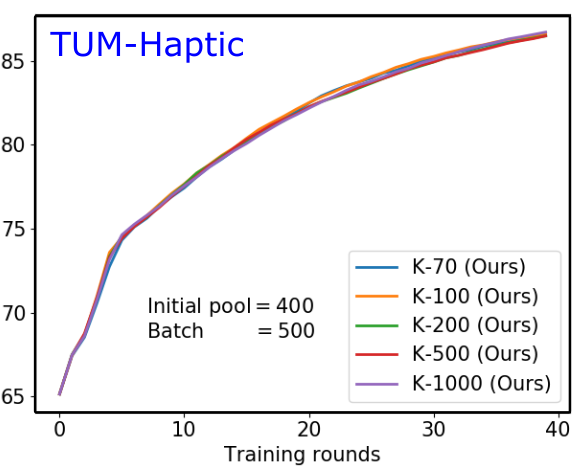}
\includegraphics[width=0.32\linewidth]{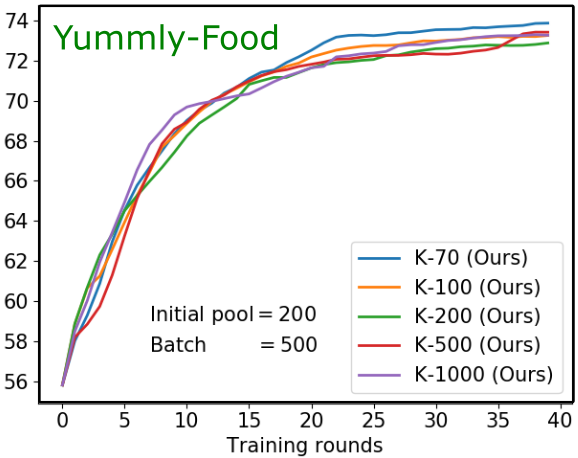} \\
\includegraphics[width=0.34\linewidth]{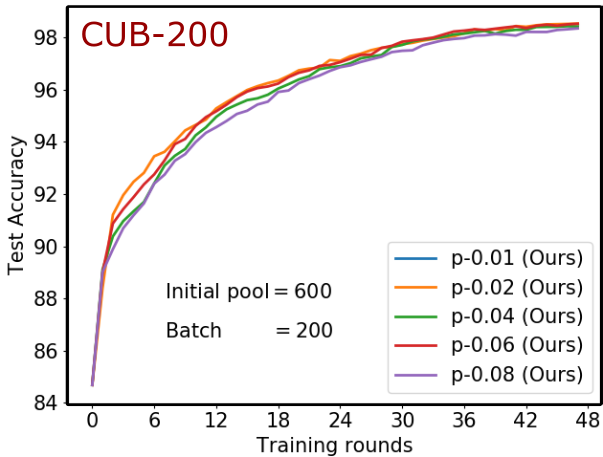}
\includegraphics[width=0.32\linewidth]{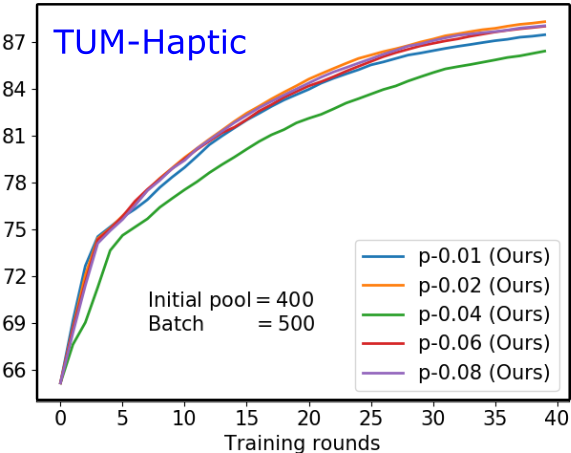}
\includegraphics[width=0.32\linewidth]{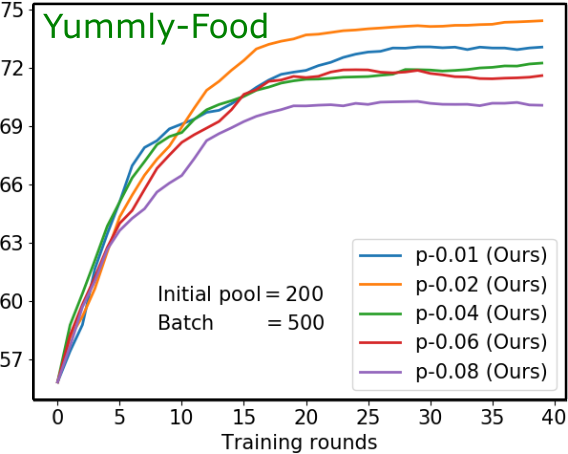}
\caption{Ablation study to analyze the robustness of our method to different values of K (\# of prior models sampled by dropout) and $p$ (dropout probability) on different datasets.\vspace{-5mm}}
\label{Q3}
\end{figure}

\paragraph{\textbf{Ablation Study}.} In order to get an estimate of the covariance matrix, we perform random dropouts in the neural network $K$ times. Naturally, as $K$ increases, one gets a better estimate of the covariance matrix. However, this may increase the computation time. We perform an ablation study to see how this hyperparameters (i.e., variation in $K$ and dropout probability $p$) affect the triplet order accuracy, and the results are shown in Figure~\ref{Q3} for three different modalities: image, haptic and food. It can be seen from the plots that a moderate value of about $K=70$ or $100$ is good enough as the performance is not significantly dependent on the choice of $K$. We also observe that the performance is robust to variation in dropout probability; however, there is a significant variation for the Yummly-Food dataset, with optimal $p=0.02$.

\paragraph{\textbf{Runtime Analysis}.} We also compare the computational requirement of the proposed method with that of Kumari \etal \cite{kumari2020batch}. The key computational step in \cite{kumari2020batch} involves searching for the subset of maximally apart (in the feature space) triplet at each training round, apart from the computation of the gradients. They also use a greedy search technique for subset selection. For the proposed method, the subset selection process is efficient, but the computation of determinant of covariance matrix at each iteration does consume a good amount of time. Overall, both the methods were found to consume a  nearly equal amount of computation time when the Gram-Schmidt orthogonalization is used. For instance runtimes (in secs) of different batch selection policies to select a 500-triplet batch from Yummly-food are: US: 0.109, Variance: 0.083, BADGE: 60.104, US-$\langle$Dist$\rangle$: 8.110, Ours: 7.803. While the computation complexity varies with the feature dimension and model size, the relative performance remains similar. 

\begin{wrapfigure}{r}{0.6\textwidth}
\centering
\vspace{-7mm}
\includegraphics[width=0.49\linewidth]{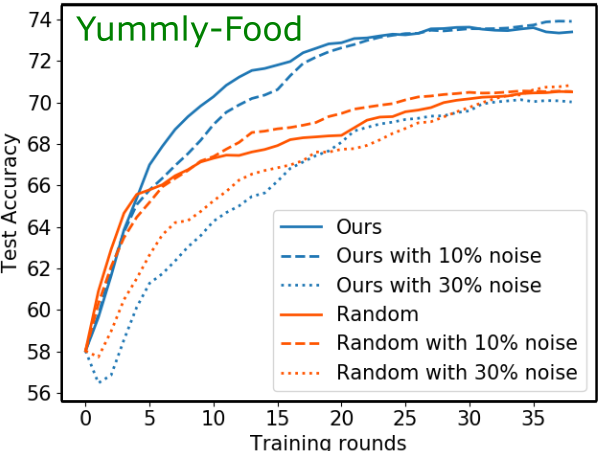}\includegraphics[width=0.49\linewidth]{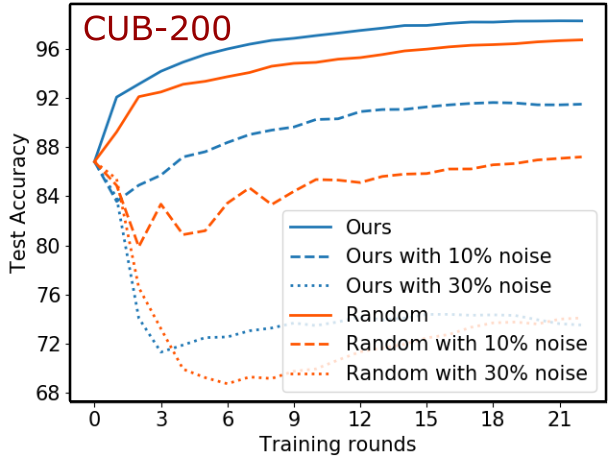}
\caption{Performance of our method vs random sampling in the presence of  labeling error.\vspace{-7mm}}
\label{Q5}
\end{wrapfigure}

\paragraph{\textbf{Robustness to Labeling Error}.} To evaluate the robustness of our method against labeling error, we corrupt $10\%$ and $30\%$ of the ground-truth training triplets in the food and image datasets by flipping their orders. Figure \ref{Q5} shows how the noisy training set affects the performance of our method vs the random sampling baseline. For the food data (top), with a relatively low $10\%$ labeling error, our method takes slightly more iterations to gain accuracy, but eventually converges to a comparably high accuracy as the noise-free case, while random batch selection fails to achieve the same performance even with clean data. As the percentage of noisy triplets increases, the performance of both methods degrade, showing vulnerability to large scale labeling error. In the absence of abnormally high levels of outliers, our method shows robust performance. For the more complex image dataset (bottom), labeling error has a stronger negative effect (the first selected batch actually decreases overall accuracy), but at each noise level our method still outperforms the baseline.

\begin{figure}[t!]
\centering
\vspace{-3mm}
\includegraphics[width=0.24\linewidth]{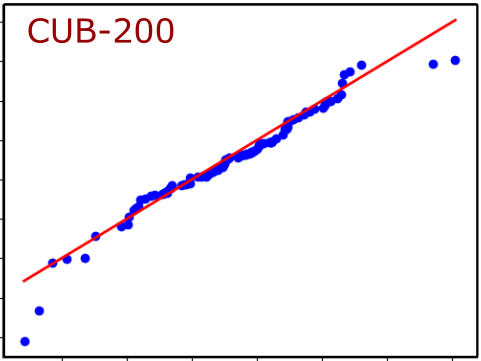}
\includegraphics[width=0.24\linewidth]{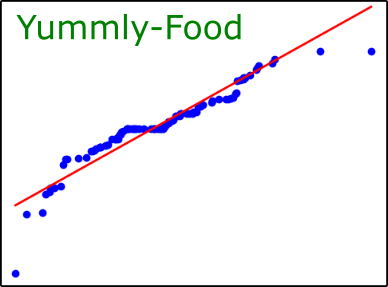}
\includegraphics[width=0.24\linewidth]{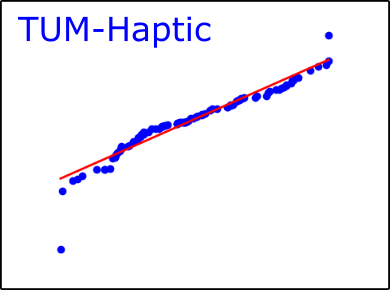}
\includegraphics[width=0.24\linewidth]{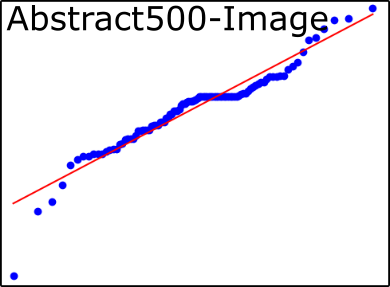}\\
\includegraphics[width=0.24\linewidth]{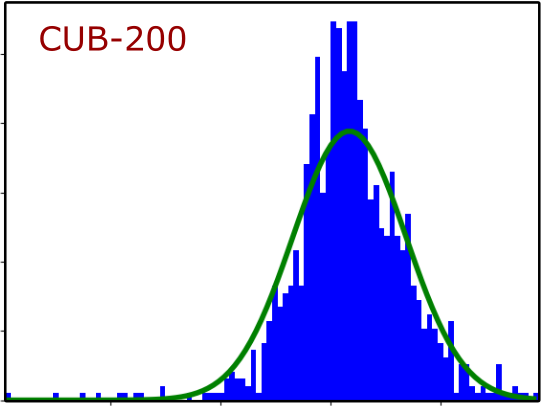}
\includegraphics[width=0.24\linewidth]{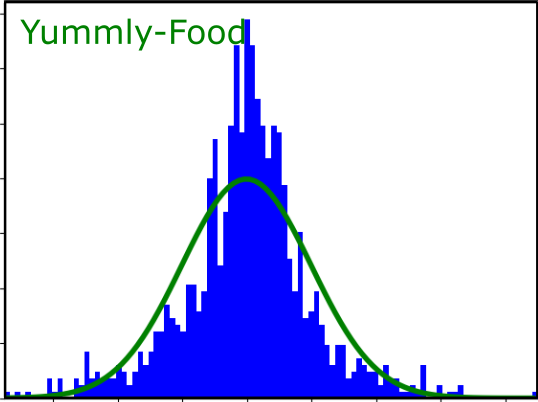}
\includegraphics[width=0.24\linewidth]{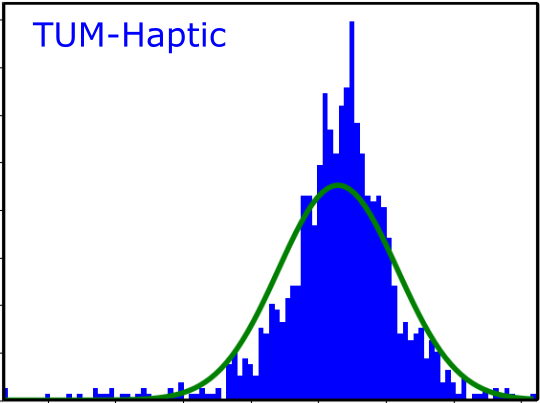}
\includegraphics[width=0.24\linewidth]{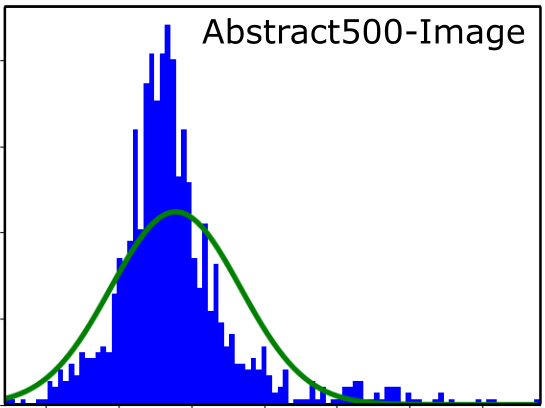}
\caption{QQ plot and histogram for all four datasets to demonstrate how closely the actual distribution follows the theoretical distribution.\vspace{-5mm}}
\label{Q4}
\end{figure}

\paragraph{\textbf{Comparison of Data Distribution to Theoretical Distribution}.} We study the validity of the Gaussian embedding, though it already has justification as ``worst-case analysis'' due to the Maximum Entropy Principle. A standard test is the quantile-quantile (QQ) plot \cite{ben2004quantile}, which indicates how close the empirical distribution is to the theoretical distribution. For ease of visualization, we show the QQ plot and histogram for a single randomly-selected unlabeled triplet, for a particular model trained on the initial triplet pool in each dataset. (We cannot visualize a full multivariate QQ plot over all possible batches.) In the QQ plot, the x-axis denotes the theoretical quantiles, which in our case is a Gaussian distribution with the empirical mean and variance, and the observed ordered distance margins are on the y-axis. The {\em goodness of fit} is indicated by the alignment of points with the straight line having a unit slope. As shown in Figure~\ref{Q4}, in all four datasets, the plotted curve closely approximates the corresponding straight lines shown in red. Our approximation is further validated in the histogram, where our data distribution shows a reasonable fit with the theoretical distribution (shown in green) for the most part, except that the actual distribution is a little more peaked.

\section{Conclusion, Limitations, and Future Work}

We have introduced a novel approach for batch-mode active metric learning based on maximizing the joint entropy of a batch. We found that a batch of individually informative triplets does not form an optimal subset, even if decorrelation heuristics are applied to reduce their correlation. Instead of defining separate measures for informativeness and diversity, our method defines the joint entropy of a batch of triplets as a unified measure that jointly optimizes both. The overall method involves no heuristic parameter selection and has no control parameter to tweak, other than the number of dropout samples and dropout probability, once the network architecture is chosen. 

While our method shows promising results, it does have a few limitations. First, approximating the joint distribution of data using the Maximum Entropy Principle gives the most general distribution for a given prior, which in the case of second-order statistics as constraints is a Gaussian. However, in some cases, where the actual distribution may be quite non-Gaussian, the joint entropy measure defined with the second-order statistics may misguide the batch selection policy. One important direction for future work is extending our framework beyond second-order statistics to learn the joint distribution of data closer to empirical distribution. Another important extension would be to modify our framework to dynamically learn the optimal batch size and batch selection policy, which we believe would further improve the performance and generalize well to diverse inputs and applications.

\bibliographystyle{splncs04}
\bibliography{egbib}

\begin{thebibliography}{10}
\providecommand{\url}[1]{\texttt{#1}}
\providecommand{\urlprefix}{URL }
\providecommand{\doi}[1]{https://doi.org/#1}

\bibitem{GNMDS}
Agarwal, S., Wills, J., Cayton, L., Lanckriet, G., Kriegman, D., Belongie, S.:
  Generalized non-metric multidimensional scaling. In: AIS (2007)

\bibitem{ash2019deep}
Ash, J.T., Zhang, C., Krishnamurthy, A., Langford, J., Agarwal, A.: Deep batch
  active learning by diverse, uncertain gradient lower bounds. In: ICLR (2020)

\bibitem{bellet2013survey}
Bellet, A., Habrard, A., Sebban, M.: A survey on metric learning for feature
  vectors and structured data. arXiv preprint arXiv:1306.6709  (2013)

\bibitem{ben2004quantile}
Ben, M.G., Yohai, V.J.: Quantile--quantile plot for deviance residuals in the
  generalized linear model. Journal of Computational and Graphical Statistics
  (2004)

\bibitem{fan2019scoot}
Fan, D.P., Zhang, S., Wu, Y.H., Liu, Y., Cheng, M.M., Ren, B., Rosin, P.L., Ji,
  R.: Scoot: A perceptual metric for facial sketches. In: ICCV (2019)

\bibitem{freytag2014selecting}
Freytag, A., Rodner, E., Denzler, J.: Selecting influential examples: Active
  learning with expected model output changes. In: ECCV (2014)

\bibitem{gal2016dropout}
Gal, Y., Ghahramani, Z.: Dropout as a bayesian approximation: Representing
  model uncertainty in deep learning. In: ICML (2016)

\bibitem{gilad2006query}
Gilad-Bachrach, R., Navot, A., Tishby, N.: Query by committee made real. In:
  NeurIPS (2006)

\bibitem{heim2015active}
Heim, E., Berger, M., Seversky, L., Hauskrecht, M.: Active perceptual
  similarity modeling with auxiliary information. In: AAAI (2015)

\bibitem{hoffmann1989iterative}
Hoffmann, W.: Iterative algorithms for gram-schmidt orthogonalization.
  Computing  (1989)

\bibitem{jamieson2011low}
Jamieson, K.G., Nowak, R.D.: Low-dimensional embedding using adaptively
  selected ordinal data. In: Allerton (2011)

\bibitem{Jaynes}
Jaynes, E.T.: Information theory and statistical mechanics. Physical Review
  (1957)

\bibitem{kendall2015bayesian}
Kendall, A., Badrinarayanan, V., Cipolla, R.: Bayesian segnet: Model
  uncertainty in deep convolutional encoder-decoder architectures for scene
  understanding. arXiv preprint arXiv:1511.02680  (2015)

\bibitem{kendall1948rank}
Kendall, M.G.: Rank correlation methods. Griffin (1948)

\bibitem{kim2019deep}
Kim, S., Seo, M., Laptev, I., Cho, M., Kwak, S.: Deep metric learning beyond
  binary supervision. In: CVPR (2019)

\bibitem{kingma2014adam}
Kingma, D.P., Ba, J.: Adam: A method for stochastic optimization. In: ICLR
  (2015)

\bibitem{kirsch2019batchbald}
Kirsch, A., van Amersfoort, J., Gal, Y.: {BatchBALD}: Efficient and diverse
  batch acquisition for deep {Bayesian} active learning. In: NeurIPS (2019)

\bibitem{kruskal1978multidimensional}
Kruskal, J.B., Wish, M.: Multidimensional scaling. Elsevier (1978)

\bibitem{priyadarshini2019perceptnet}
Kumari, P., Chaudhuri, S., Chaudhuri, S.: {PerceptNet}: Learning perceptual
  similarity of haptic textures in presence of unorderable triplets. In: WHC
  (2019)

\bibitem{kumari2020batch}
Kumari, P., Goru, R., Chaudhuri, S., Chaudhuri, S.: Batch decorrelation for
  active metric learning. In: IJCAI-PRICAI (2020)

\bibitem{mcfee2011learning}
McFee, B., Lanckriet, G.: Learning multi-modal similarity. JMLR  (2011)

\bibitem{nemhauser1978analysis}
Nemhauser, G.L., Wolsey, L.A., Fisher, M.L.: An analysis of approximations for
  maximizing submodular set functions -- {I}. Mathematical programming  (1978)

\bibitem{oliva2001modeling}
Oliva, A., Torralba, A.: Modeling the shape of the scene: A holistic
  representation of the spatial envelope. IJCV  (2001)

\bibitem{pinsler2019bayesian}
Pinsler, R., Gordon, J., Nalisnick, E., Hern{\'a}ndez-Lobato, J.M.: Bayesian
  batch active learning as sparse subset approximation. In: NeurIPS (2019)

\bibitem{robb2015crowdsourced}
Robb, D.A., Padilla, S., Kalkreuter, B., J.Chantler, M.: Crowdsourced feedback
  with imageryrather than text: Would designers use it? In: SIGCHI (2015)

\bibitem{CNN2018ICLR}
Sener, O., Savarese, S.: Active learning for convolutional neural networks: A
  core-set approach. In: ICLR (2018)

\bibitem{settles2012}
Settles, B.: Active learning. SLAIML  (2012)

\bibitem{shui2020deep}
Shui, C., Zhou, F., Gagn{\'e}, C., Wang, B.: Deep active learning: Unified and
  principled method for query and training. In: AISTATS (2020)

\bibitem{sinha2019variational}
Sinha, S., Ebrahimi, S., Darrell, T.: Variational adversarial active learning.
  In: ICCV (2019)

\bibitem{Strese}
Strese, M., Boeck, Y., Steinbach, E.: Content-based surface material retrieval.
  In: WHC (2017)

\bibitem{tamuz2011adaptively}
Tamuz, O., Liu, C., Belongie, S., Shamir, O., Kalai, A.T.: Adaptively learning
  the crowd kernel. In: ICML (2011)

\bibitem{wah2015learning}
Wah, C., Maji, S., Belongie, S.: Learning localized perceptual similarity
  metrics for interactive categorization. In: WACV (2015)

\bibitem{wang2014learning}
Wang, J., Song, Y., Leung, T., Rosenberg, C., Wang, J., Philbin, J., Chen, B.,
  Wu, Y.: Learning fine-grained image similarity with deep ranking. In: CVPR
  (2014)

\bibitem{wilber2014cost}
Wilber, M.J., Kwak, I.S., Belongie, S.J.: Cost-effective hits for relative
  similarity comparisons. In: AAAI (2014)

\bibitem{xing2003distance}
Xing, E.P., Jordan, M.I., Russell, S.J., Ng, A.Y.: Distance metric learning
  with application to clustering with side-information. In: NeurIPS (2003)

\bibitem{perceptualMetric}
Zhang, R., Isola, P., Efros, A.A., Shechtman, E., Wang, O.: The unreasonable
  effectiveness of deep features as a perceptual metric. In: CVPR (2018)

\end{thebibliography}

 \title{A Unified Batch Selection Policy for Active Metric Learning}
 \author{Supplementary Material}
 \institute{~}

 \maketitle

In this supplementary material, we provide (a) plots with standard deviations for three datasets where only the mean accuracies were provided in the main paper, for clarity; (b) performance evaluation of different active learning methods on one additional dataset and (c) qualitative and quantitative results on a retrieval task with metrics trained with different methods.

\section{Standard Deviation Plots}

\begin{figure*}[!hb]
\centering
\begin{tabular}{ccc}
\includegraphics[width=0.327\linewidth]{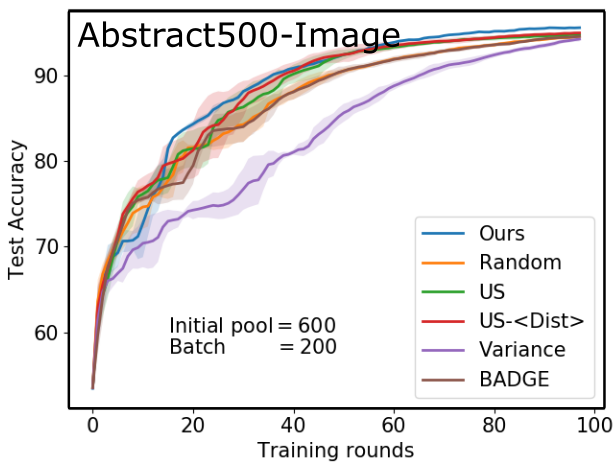}
\includegraphics[width=0.327\linewidth]{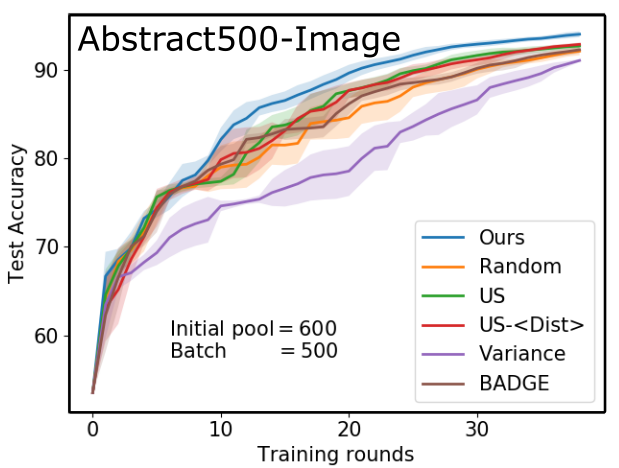}
\includegraphics[width=0.327\linewidth]{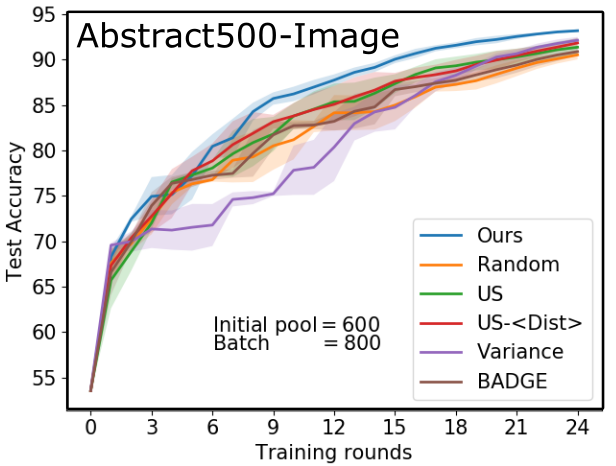}\\
\includegraphics[width=0.327\linewidth]{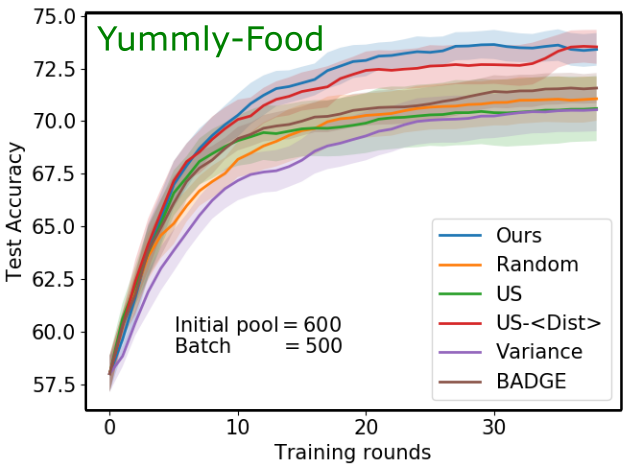}
\includegraphics[width=0.327\linewidth]{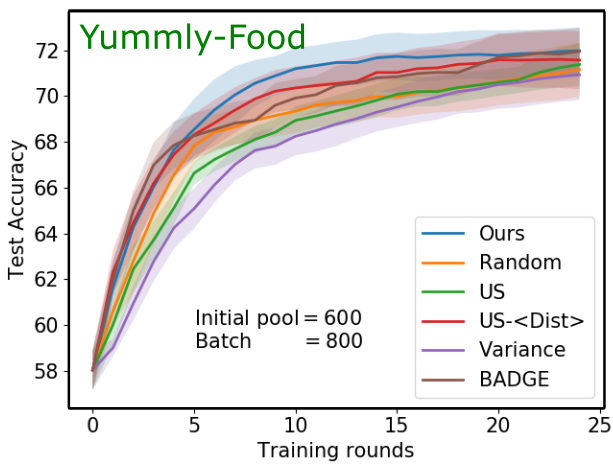}
\includegraphics[width=0.327\linewidth]{Results/food_init_600_inr_800test_acc_var.png}\\
\includegraphics[width=0.327\linewidth]{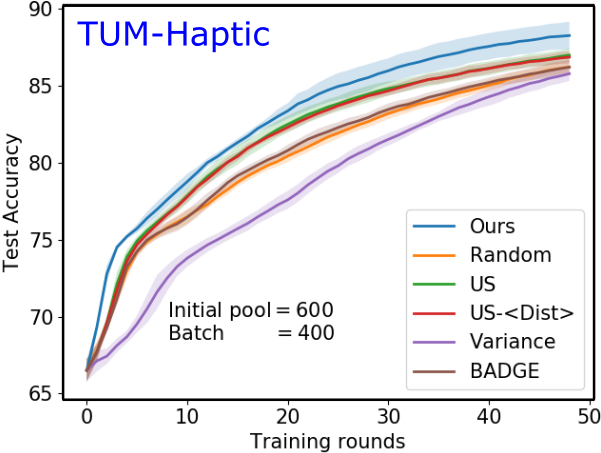}
\includegraphics[width=0.327\linewidth]{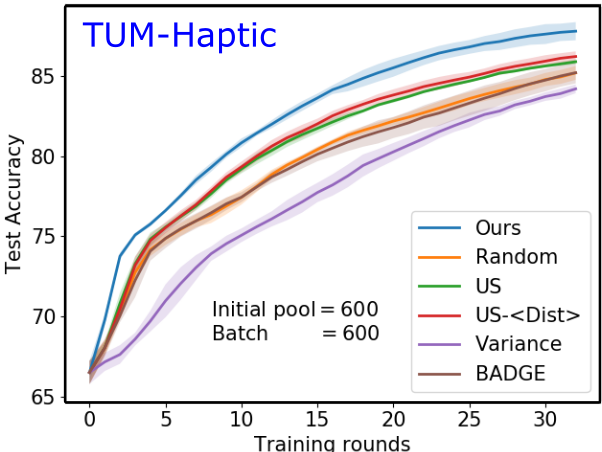}
\includegraphics[width=0.327\linewidth]{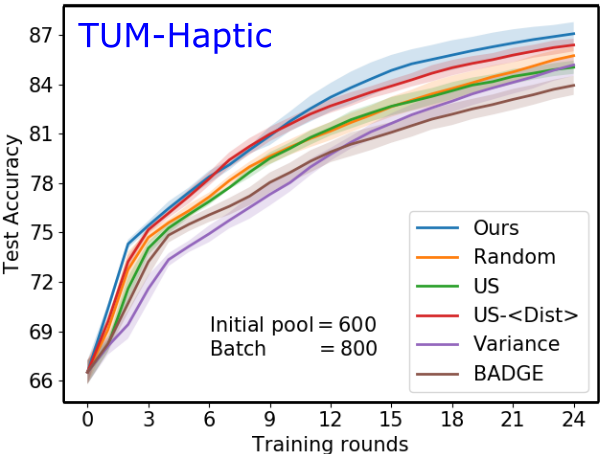}\\
\end{tabular}
\caption{Mean accuracy and standard deviation plots (computed over five random train/test splits) of different active learning methods for Abstract500-Image, Yummly-Food and TUM-Haptic datasets with increasing batch sizes.}
\label{S1}
\end{figure*}

In Figure~\ref{S1}, we augment the plots shown in the main paper with standard error bands across the five experimental runs for each method+hyperparameter combination for three real-world datasets: Abstract500-Image, Yummly-Food, and TUM-Haptic. (Bands for CUB-200 are directly shown in the main paper.) The standard error of our method is small for all datasets except the Yummly-Food dataset, where all methods have high variance. While our method always exceeds or matches the best alternative, the performance gain is observed to be higher for a larger batch size.

\section{Performance of Active Learning Methods on Scoot Facial Sketch Dataset}

The Scoot facial sketch dataset is relatively small, consisting of just 1282 triplets, where each triplet represents similarity ordering between three sketched faces of a person. The facial sketch is represented by a 512-D GIST feature extracted using 32 Gabor filters at four scales and eight orientations. The training and test set contains 800 and 200 triplets, respectively, sampled from the entire triplet set. The architecture and training hyperparameters used for Scoot facial sketch dataset are: 6 FC layers with 512, 256, 128, 64, 32 and 16 neurons. Each layer is followed by a dropout layer with a dropout probability of 0.02. The Adam optimizer is used for training the model with a learning rate of $10^{-6}$. The model is trained with an SGD batch size of 500 for 1000 epochs.

\begin{figure*}[!b]
\centering
\includegraphics[width=0.6\linewidth]{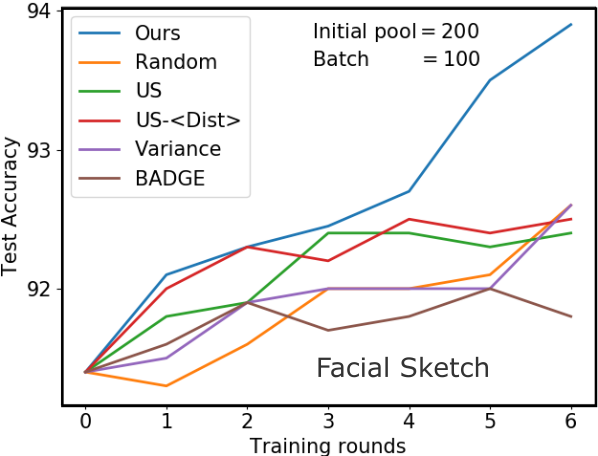}
\caption{Performance of different active learning methods on the Scoot dataset. Test accuracy indicates the fraction of test triplets correctly ordered by the learned perceptual metric.}
\label{S2}
\end{figure*}

As can be seen in Figure~\ref{S2}, in the initial training rounds with only a few annotated triplets, our method performs marginally better than other baselines, but the accuracy margin improves as the number of annotated triplets increases. 

\section{Image Retrieval Task}

In this section, we further evaluate the effectiveness of our method for an image retrieval task. We compare our method with the random sampling baseline at different training rounds on the CUB-200 and Yummly-Food datasets.

\begin{figure*}[!t]
\centering
\begin{tabular}{ccc}
\includegraphics[width=0.5\linewidth]{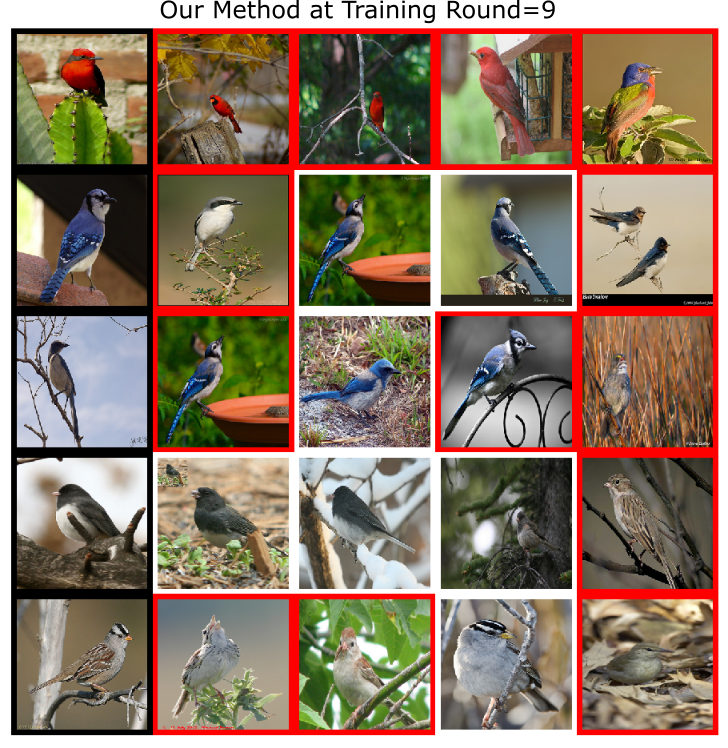} \hspace{3mm}
\includegraphics[width=0.5\linewidth]{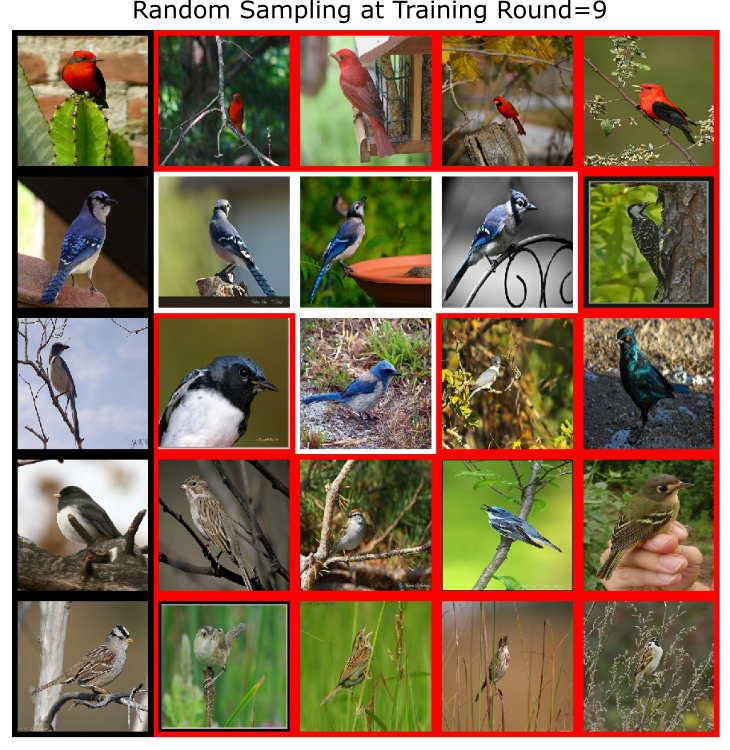}\\
\includegraphics[width=0.5\linewidth]{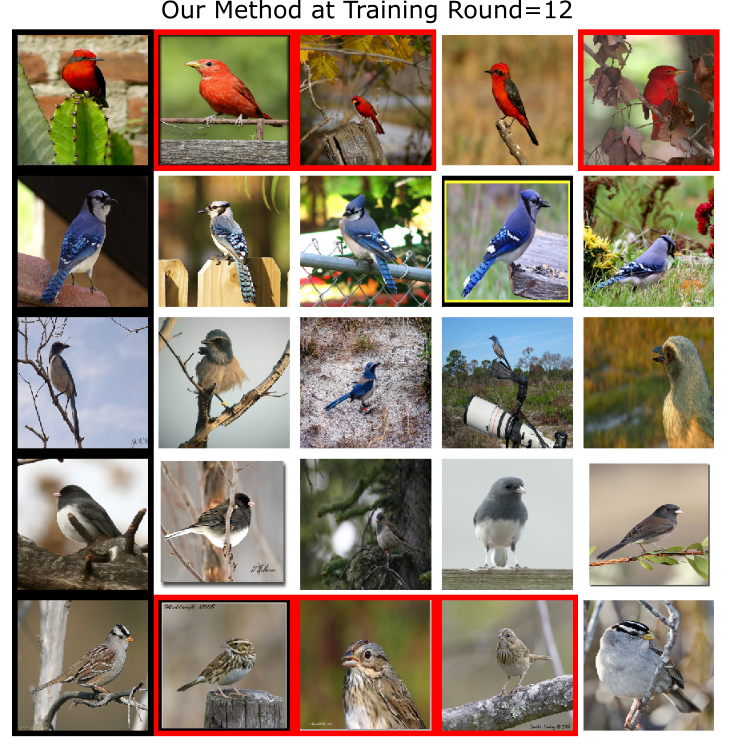} \hspace{3mm}
\includegraphics[width=0.5\linewidth]{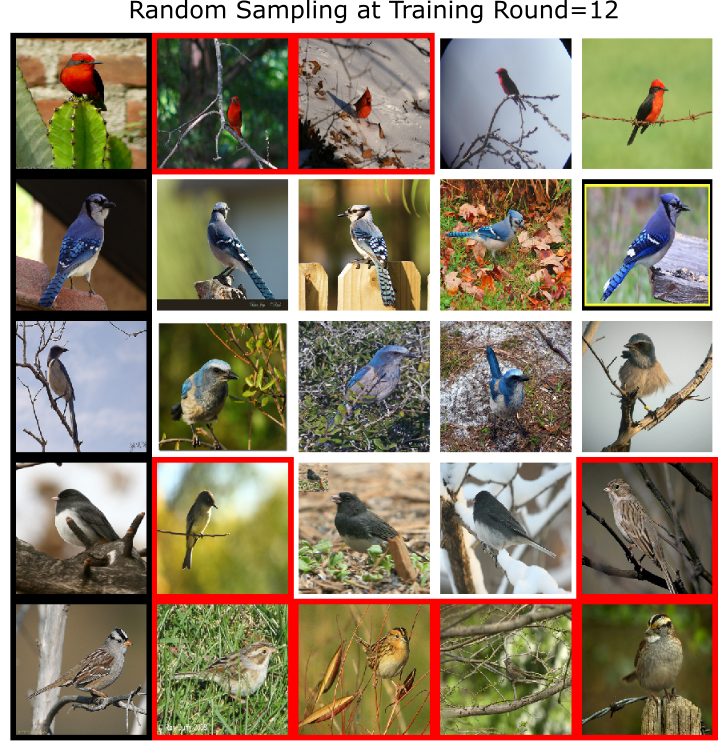}\\
\end{tabular}
\caption{Performance comparison between our method and random sampling for the image retrieval task after two different training rounds. The leftmost column presents a query, and the remaining columns present the first four retrieved images in the order of increasing perceptual distance, left to right. Images with red squares belong to classes different from the query class. Our results indicate that two images from different classes can be visually more similar than two from the same class, highlighting the distinction between perceptual and class-based metrics. The triplet ordering accuracy for $M^k_{Ours/Random}$ (model trained on triplets selected by our method vs random sampling resp. at $k^\text{th}$ training round): $M^9_{\text{Ours}} = 95.3\%$, $M^9_{\text{Random}} = 90.3\%$, $M^{12}_{\text{Ours}} = 96.7\%$, $M^{12}_{\text{Random}} = 92\%$.}
\label{S3}
\end{figure*}

\subsection{Retrieval Results on CUB-200 Dataset}

\begin{figure*}[!ht]
\centering
\begin{tabular}{ccc}
\includegraphics[width=0.5\linewidth]{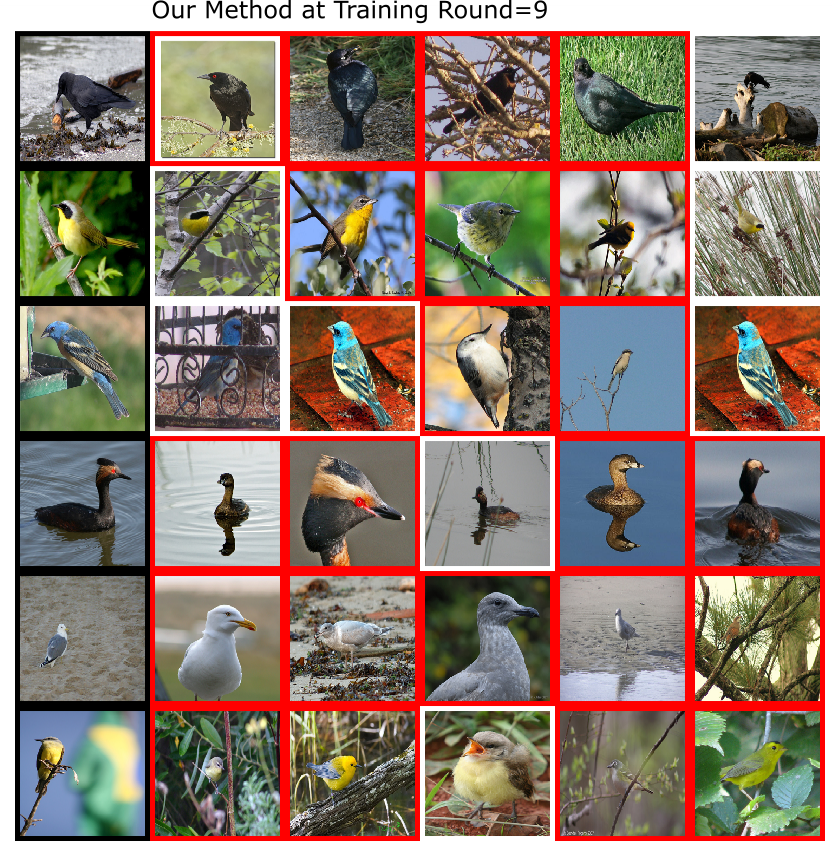} \hspace{3mm}
\includegraphics[width=0.5\linewidth]{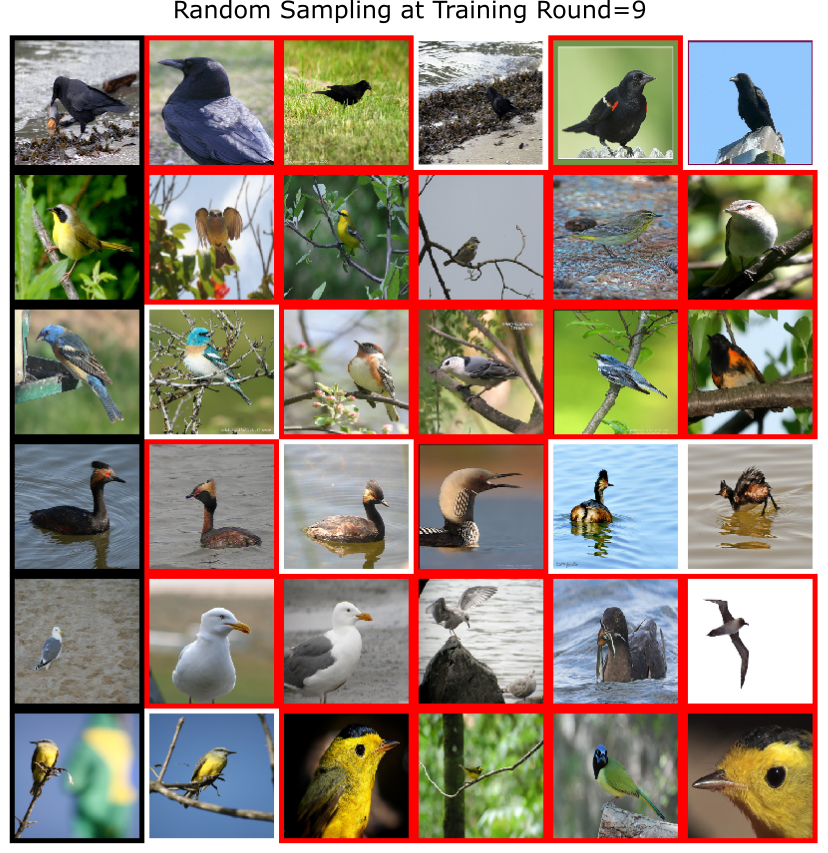}\\
\includegraphics[width=0.5\linewidth]{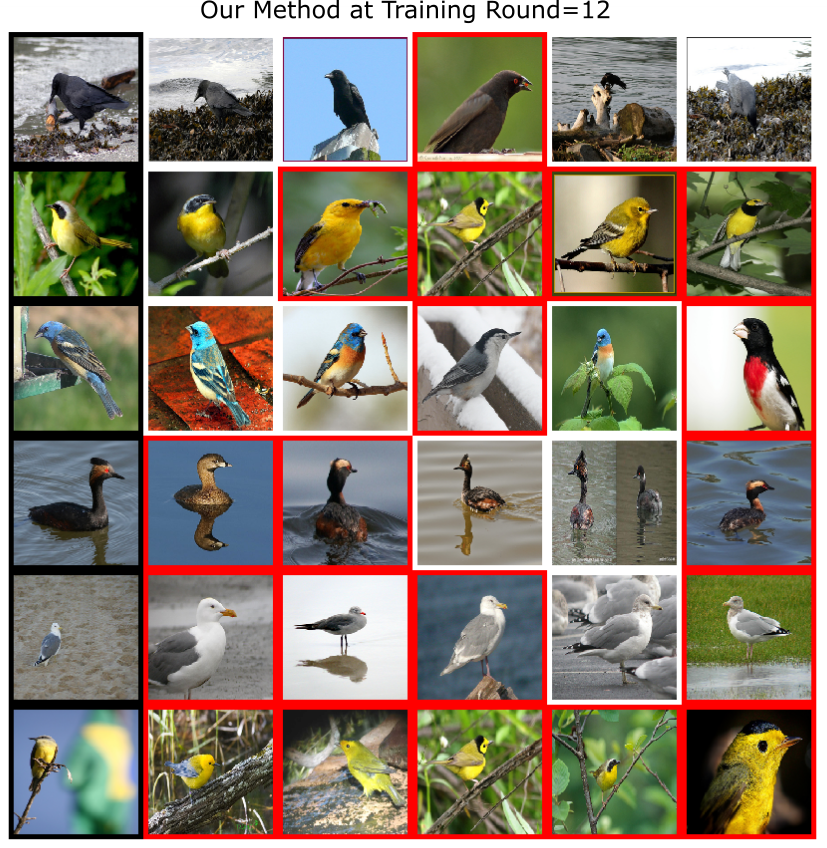} \hspace{3mm}
\includegraphics[width=0.5\linewidth]{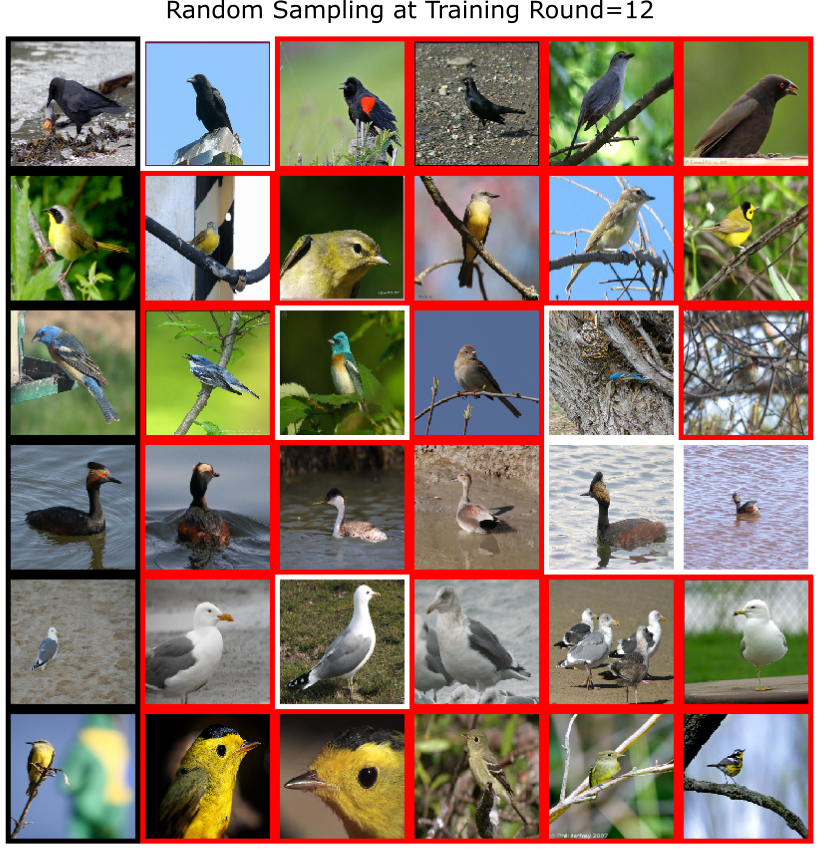}\\
\end{tabular}
\caption{Similar results as Figure~\ref{S3}, shown for a different set of query images in the leftmost column.}
\label{S4}
\end{figure*}

The CUB-200 dataset consists of 200 classes with roughly 30 images in each class. Triplets are defined based on visual (perceptual) similarity of classes. We split the dataset into 40000 training and 33000 test triplets, and performed active learning with a batch size of 600 and an initial pool of 500 triplets. For a given query image, the top four instances from the retrieval set are shown (ranked from most similar to least similar) in Figures~\ref{S3} and \ref{S4}, after two different training rounds.

\subsection{Retrieval Results on Yummly-Food Dataset}
\begin{figure*}[!t]
\centering
\begin{tabular}{ccc}
\subfloat[Retrieval results of our method using annotations of 18\% of training  triplets.]{\includegraphics[width=1.2\linewidth]{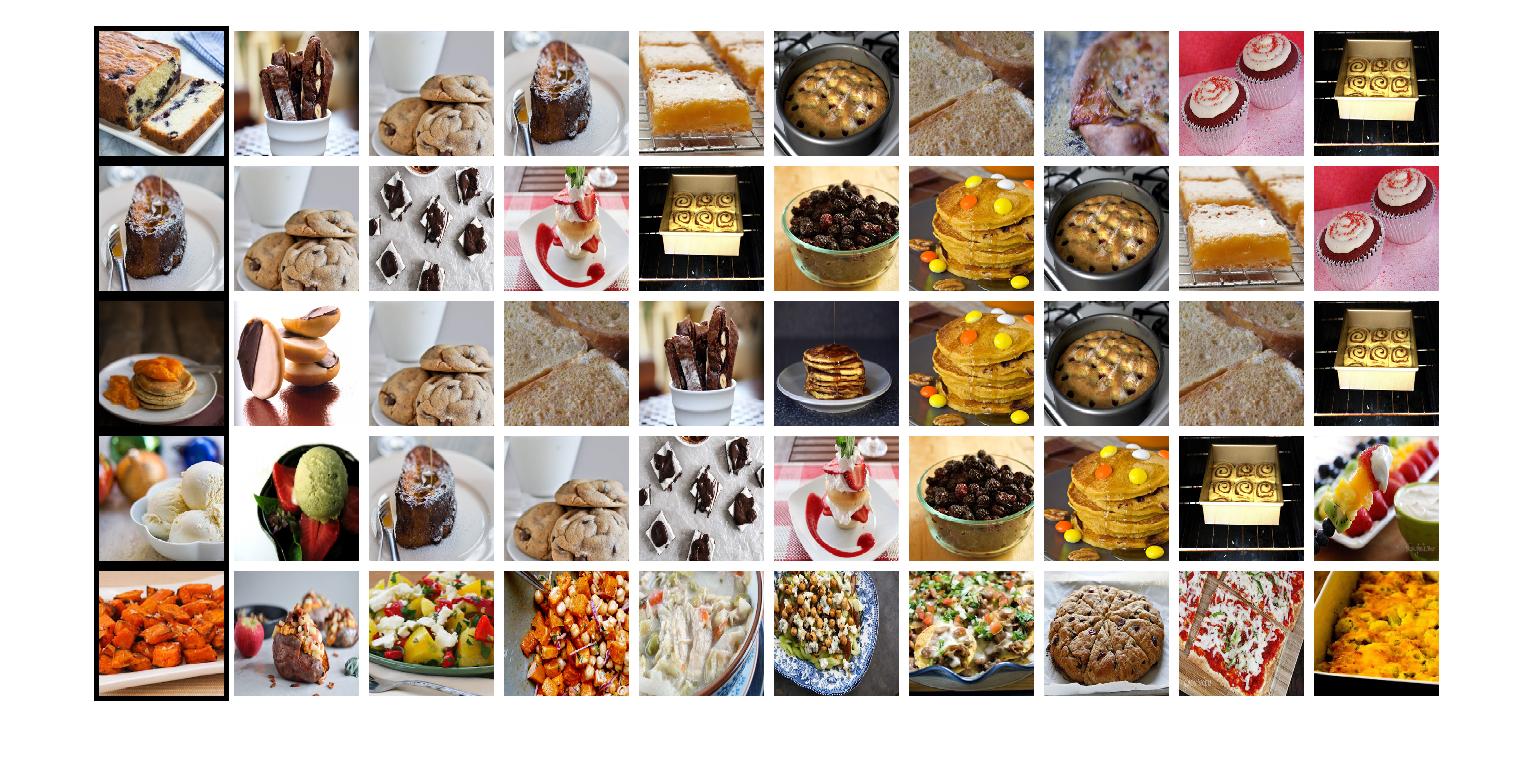}}\\
\subfloat[Retrieval results of the random sampling baseline using annotations of 18\% of training triplets.]{\includegraphics[width=1.2\linewidth]{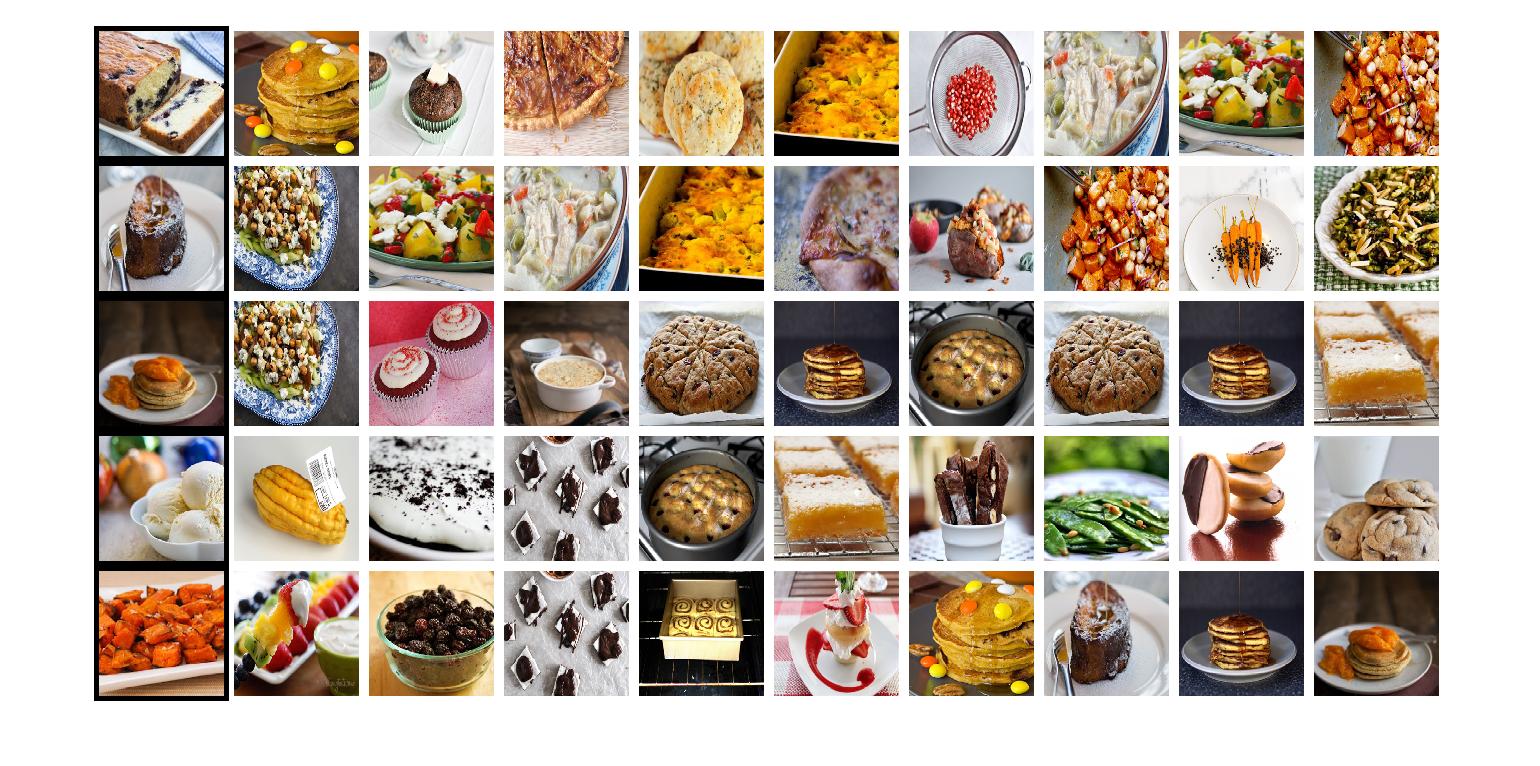}}\\
\end{tabular}
\caption{Top-9 food dishes, retrieved according to taste similarity, by our method vs random sampling after twelve training rounds, for query images in the leftmost columns. The triplet ordering accuracy for $M^k_{Ours/Random}$ (model trained on triplets selected by our method vs random sampling resp. at $k^\text{th}$ training round): $M^{12}_{\text{Ours}} = 72.9\%$, $M^{12}_{\text{Random}} = 69.3\%$. Note that both our method and random sampling solicit annotations of 18\% of the training triplets; however, our method's retrieval results better resemble the query in taste, than those from random sampling.}
\label{S5}
\end{figure*}

\begin{figure*}[!ht]
\centering
\begin{tabular}{ccc}
\subfloat[Retrieval results of our method using annotations of 30\% of training triplets.]{\includegraphics[width=1.2\linewidth]{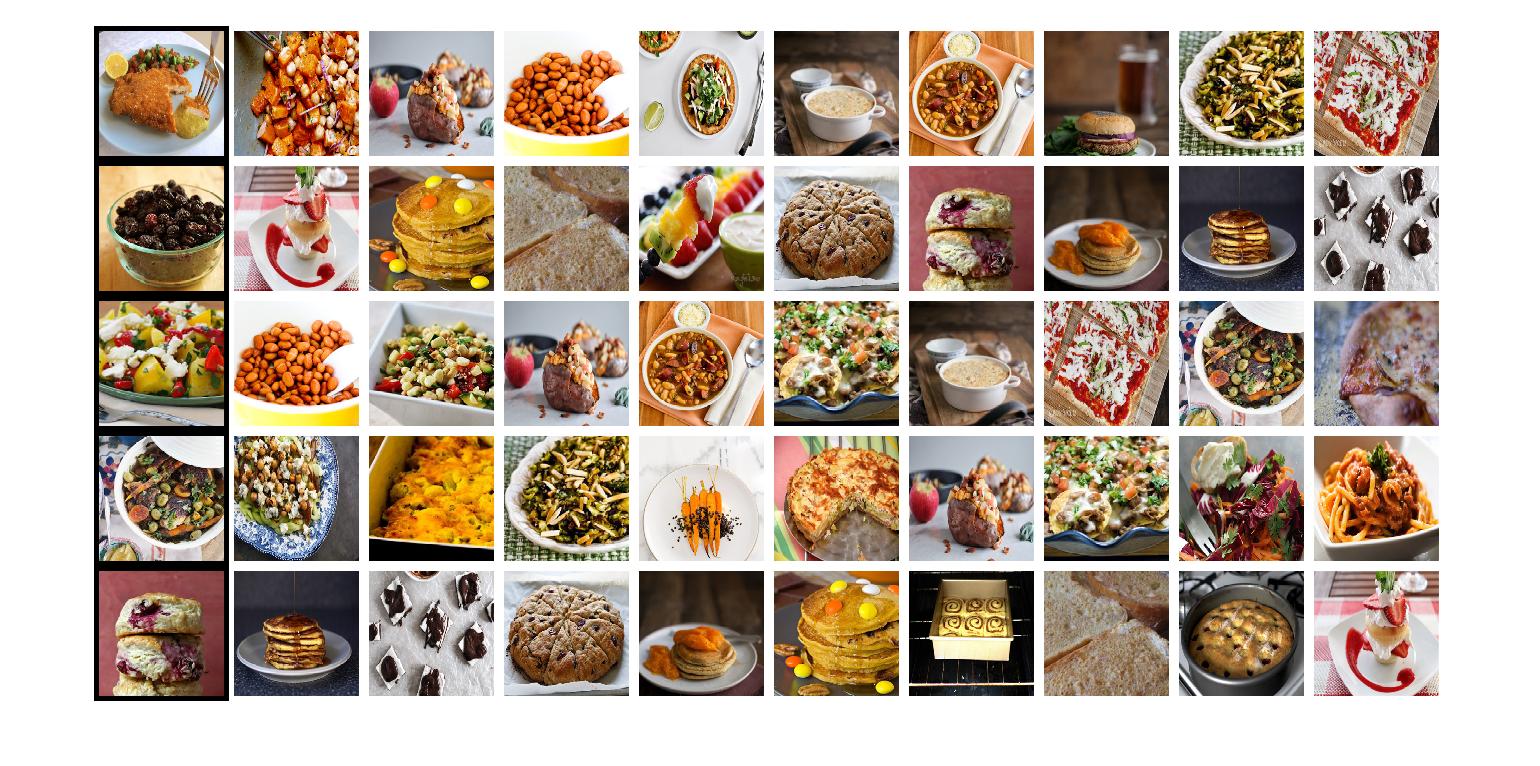}}\\
\subfloat[Retrieval results of the random sampling baseline using annotations of 30\% of training triplets.]{\includegraphics[width=1.2\linewidth]{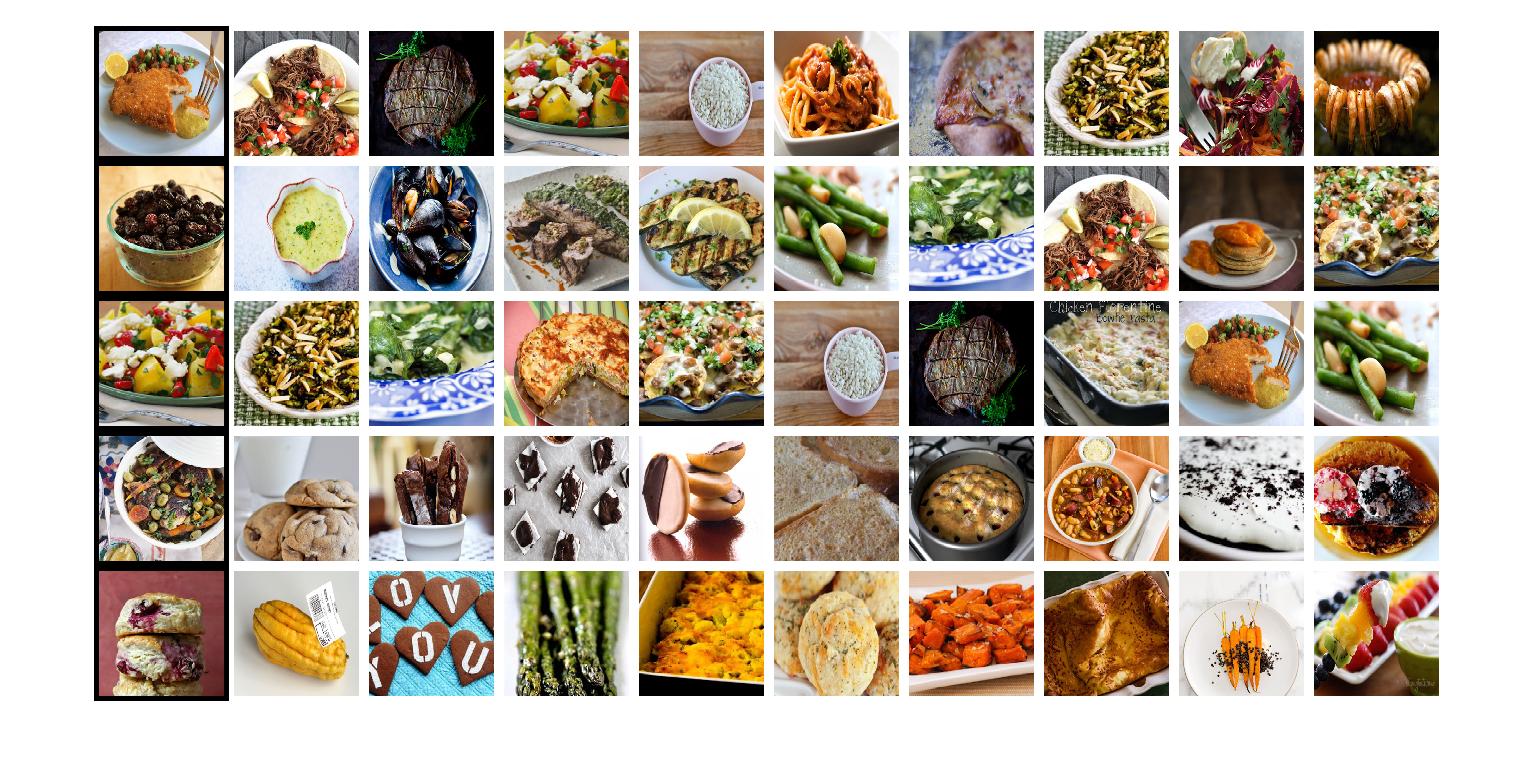}}\\
\end{tabular}
\caption{Top-9 retrieved food dishes by our method vs random sampling after twenty training rounds, for query images in the leftmost columns. The triplet ordering accuracy: $M^{20}_{\text{Ours}} = 73.6\%$, $M^{20}_{\text{Random}} = 69.7\%$. }
\label{S6}
\end{figure*}

The Yummly-Food dataset consists of images of 73 food items and 72148 triplets defined on the basis of similarity in taste. We split the dataset into 40000 training and 32148 test triplets, and performed active learning with a batch size of 600 and an initial pool of 500 triplets. For a given query image, the top nine instances from the retrieval set are shown (ranked from most similar to least similar) in Figures~\ref{S5} (9 training rounds) and \ref{S6} (20 training rounds).

\end{document}